\documentclass[11pt]{article}

\usepackage[preprint]{acl}

\usepackage{times}
\usepackage{latexsym}

\usepackage{float}

\usepackage[T1]{fontenc}
\usepackage[utf8]{inputenc}

\usepackage{booktabs}
\usepackage{hyperref}
\usepackage{amssymb}
\usepackage{amsmath}
\usepackage[nameinlink,noabbrev]{cleveref}

\newcommand{\appref}[1]{\hyperref[#1]{Appendix~\ref*{#1}}}

\usepackage{microtype}

\usepackage{inconsolata}

\usepackage{graphicx}

\title{A Geometric Account of Activation Steering through Angle--Norm Decomposition}

\author{Georgii Aparin \\
  Huawei Noah's Ark Lab \\
  \texttt{aparingm@gmail.com} \\\And
  Tatiana Gaintseva \\
  Queen Mary University of London \\
  \texttt{t.gaintseva@qmul.ac.uk}
}

\begin{document}
\maketitle
\begin{abstract}
Linear activation steering has gained popularity as a simple and empirically effective way to control language model behavior. More recently, spherical steering paradigms have been proposed to address limitations of additive interventions, often motivated by the assumption that hidden-state norm does not carry concept-relevant information. In this work, we revisit this assumption through a controlled empirical study designed to disentangle the roles of angular and radial components. We show that steering methods differ mainly in how they couple two geometric effects: changing a token's angular alignment with a concept direction and changing its hidden-state norm. Across seven language models, we find that concepts are represented primarily in angular structure, supporting the motivation for spherical methods, but that norm remains important for the stability and downstream effects of steering. Our results explain why interventions with similar concept-level effects can behave differently, and suggest that activation steering should be parameterized by interpretable angular and radial components of the intervention, rather than by a single additive coefficient that entangles these two effects.
\end{abstract}

\section{Introduction}

Linear activation steering has become a widely used approach for controlling language model behavior through interventions on intermediate representations~\citep{zou2023representation, turner2023activation, panickssery2023steering}. Given a steering direction associated with a target concept, standard methods add this direction to hidden states with a scalar strength. These interventions are simple, training-free, and effective across behaviors such as truthfulness, sentiment, toxicity, and refusal~\citep{zou2023representation, turner2023activation, panickssery2023steering, li2023inference, rimsky2024steering, arditi2024refusal}. However, additive steering treats activation space as if concept control were naturally linear: increasing the steering coefficient is assumed to move representations in a meaningful behavioral direction. This obscures the geometry of the intervention, since adding a vector changes both the direction and the norm of the hidden state~\citep{park2024linear, vu2025angular, you2026spherical}.

\begin{figure}[t]
\centering
\includegraphics[width=\linewidth]{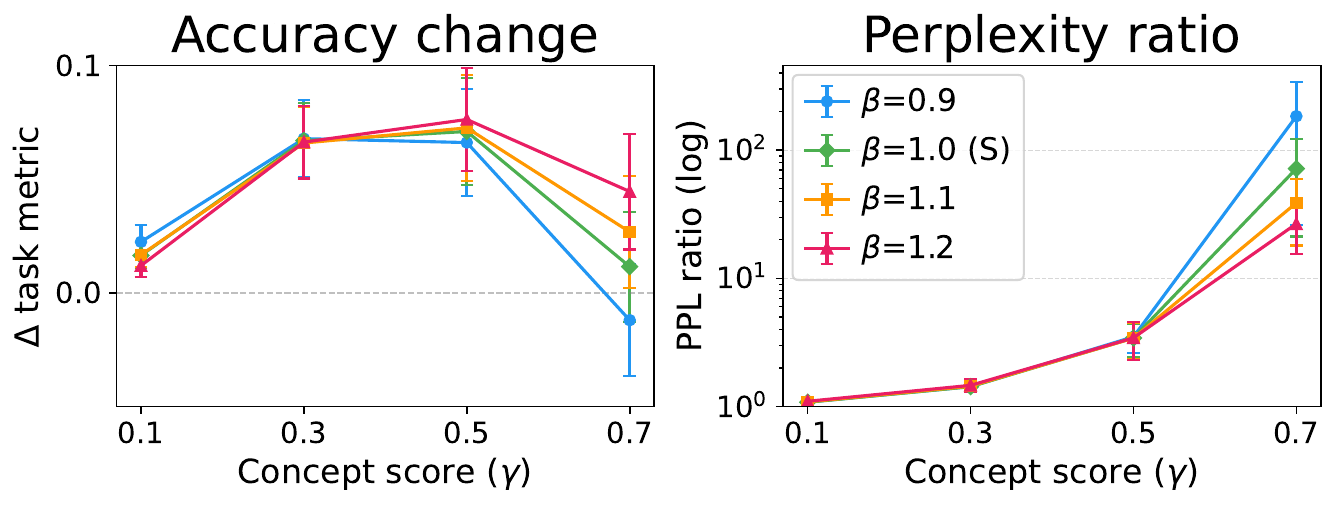}
\caption{
Effect of norm scaling in SN. The left panel shows downstream task metric change, and the right
panel shows perplexity ratio. Increasing $\beta$ has little effect on the semantic task metric but substantially reduces perplexity at high $\gamma$, indicating that the norm primarily controls generation stability.
}
\label{fig:beta_effect}
\end{figure}

\begin{figure}[t]
\centering
\includegraphics[width=\linewidth]{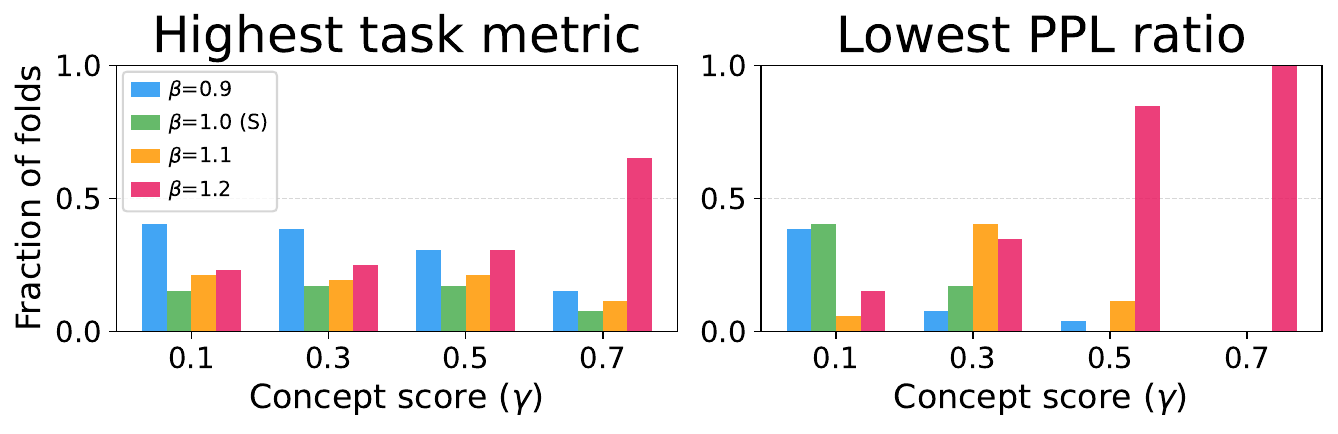}
\caption{
Fraction of folds in which each $\beta$ value achieves the best perplexity or task metric. At
$\gamma=0.7$, $\beta=1.2$ achieves the lowest perplexity in all folds in our evaluation, indicating that strict norm preservation is not always the most stable choice for high-strength spherical steering.
}
\label{fig:beta_wins}
\end{figure}

Recent angular and spherical steering methods offer an alternative: instead of translating activations, they rotate hidden states toward a concept direction, often while preserving norm~\citep{vu2025angular, you2026spherical}. This is motivated by the hypothesis that concept information is primarily angular, while norm preservation maintains generation quality and input relevance. Although spherical methods can improve stability over naive additive steering, their underlying assumptions remain insufficiently examined: do concepts mainly live in activation direction, and is strict norm preservation always the right steering constraint?

We study these questions through a controlled geometric comparison of activation steering methods. We decompose each hidden state into an angular component, which determines alignment with a concept direction, and a radial component, given by its norm. This lets us compare six steering approaches that differ in whether they enforce a target angular concept score, preserve the original norm, or allow the norm to change.

Our experiments show that the angular hypothesis is largely correct. Across seven language models and four concept datasets, probes trained on normalized hidden states closely match probes trained on raw hidden states, while norm-only probes remain near chance. Thus, for the concepts we study, concept-discriminative information is encoded primarily in activation direction rather than magnitude.

However, norm is not irrelevant. Although activation magnitude does not directly encode the target concept, it plays an important role in generation stability and capability preservation. At high angular strength, strict norm preservation can cause large perplexity increases and capability degradation. Conversely, methods that reach the same angular target while allowing a modest norm increase often better preserve fluency and downstream performance (Figs.~\ref{fig:beta_effect},~\ref{fig:beta_wins}). This yields a more nuanced conclusion than either additive or spherical steering alone suggests: angular control explains semantic steering, but radial scaling can determine whether the intervention remains usable at high strength.

We hypothesize that hidden-state norm partly controls the effective representational capacity available at a token. Under strong steering, forcing a target concept into the original fixed radius may leave less capacity for other context-relevant information. A modest norm increase can relieve this pressure, allowing the model to express the desired concept direction while retaining enough representational scale for other features.

Overall, our findings suggest that activation steering should be viewed neither as a one-parameter additive intervention nor as a purely angular operation with fixed norm. Instead, steering is better understood as a two-parameter geometric intervention governed by both angle and radius: angle controls the intended semantic effect, while radius influences generation stability, input relevance, and capability preservation. This perspective explains why methods with similar concept-level effects can behave differently and suggests a more interpretable design space for future steering methods.

Our contributions are as follows:
\begin{itemize}
    \item We \textbf{formulate activation steering as a two-component geometric intervention} that separates angular concept control from radial norm modification.
    \item We \textbf{compare six steering methods under a common framework}, distinguishing whether they preserve norm and whether they enforce a per-token target concept score.
    \item We \textbf{empirically test the angular encoding hypothesis} across seven language models and four concept datasets, finding that concept information is primarily encoded in activation direction.
    \item We \textbf{show that norm still plays a crucial role in steering stability}: at high steering strengths, modest norm increases can reduce perplexity by up to $1.8\times$ without substantially changing the semantic steering effect.
\end{itemize}
\section{Related Work}

\noindent \textbf{Activation steering and representation engineering.}
Activation steering controls model behavior by modifying intermediate activations at inference time, without updating weights. Most methods identify a direction in hidden-state space associated with a target behavior and intervene along this direction during generation. ITI, ActAdd, CAA, and Representation Engineering have been used to affect truthfulness, sentiment, topic, refusal, toxicity, and other high-level attributes~\citep{li2023inference, turner2023activation, panickssery2023steering, zou2023representation, rimsky2024steering, arditi2024refusal}. These methods are simple and training-free, but their usual additive strength has unclear geometry: changing it alters both the hidden state's alignment with the steering direction and its norm. Our work studies this ambiguity directly by decomposing steering into angular and radial components.

\smallskip
\noindent \textbf{Linear concept representations.}
Activation steering is closely related to the hypothesis that high-level model properties are represented linearly in activation space. Under this view, directions in hidden-state space correspond to concepts or behaviors, and projections onto these directions can serve as concept scores~\citep{park2024linear, zou2023representation}. This motivates contrastive direction extraction, probing, and direction-based interventions. However, identifying a useful concept direction does not determine how to intervene along it. Additive steering simultaneously changes angular alignment and representation norm, which may play different roles. We therefore separate two often conflated questions: whether concept information is encoded in activation direction, and how norm changes affect steering outcomes.

\smallskip
\noindent \textbf{Angular and spherical steering.}
Recent work has proposed angular or spherical alternatives to additive steering. Angular Steering rotates activations in a behavior-related subspace~\citep{vu2025angular}, while Spherical Steering performs a norm-preserving geodesic rotation toward a target direction~\citep{you2026spherical}. These methods are motivated by the idea that concept information is primarily angular and that preserving activation norm helps maintain generation quality. Our work provides a controlled examination of these assumptions: we test whether concepts are indeed primarily encoded in direction, and whether strict norm preservation remains desirable once angular control is fixed. Unlike prior work focused on specific steering rules, we analyze additive, renormalized, matched, angular, spherical, and norm-scaled interventions in a single angular--radial framework.

\smallskip
\noindent \textbf{Adaptive and token-wise steering.}
Several methods suggest that a single global steering coefficient is insufficient. ITI selects intervention sites at the attention-head level~\citep{li2023inference}; Representation Engineering studies control directions across layers and behaviors~\citep{zou2023representation}; and selective or adaptive steering methods vary interventions across layers, tokens, or examples to reduce side effects~\citep{dang2026selective}. Our results clarify what such adaptation should control: the achieved angular concept score and the radial norm scale. This yields a more interpretable design space in which methods differ not only in average steering strength, but also in per-token angular precision and norm handling.

\smallskip
\noindent \textbf{Our contribution.}
Prior work shows that activation directions can control behavior, and recent spherical methods show that norm-aware interventions can improve stability. We ask which geometric component is responsible for concept control, and which affects stability. Our experiments indicate that the evaluated concepts are primarily encoded in activation direction, supporting the motivation for spherical steering. At the same time, norm is not merely nuisance variation: modest radial changes can substantially affect perplexity and capability preservation even when angular concept score is fixed. Thus, we reframe activation steering as a two-parameter intervention over angle and radius, rather than a one-dimensional choice of additive strength or a binary choice between additive and norm-preserving methods.

\section{Methodology}

We study activation steering as a geometric intervention on the residual-stream hidden state of a language model. Given a hidden state $x \in \mathbb{R}^d$ at a fixed transformer layer and a unit steering direction $s$, we decompose $x$ into a radial component and an angular component:
\begin{equation}
    r = \|x\|, \qquad
    u = \frac{x}{r},
    \label{eq:radial-angular-decomposition}
\end{equation}
\begin{equation}
    c = \langle u, s\rangle, \qquad
    v = \frac{u - c s}{\|u - c s\|}.
    \label{eq:concept-residual-decomposition}
\end{equation}
Here, $r$ is the hidden-state norm, $u$ is the corresponding unit vector, $c$ is the angular concept score, and $v$ is the unit residual direction orthogonal to $s$. Any unit vector in the two-dimensional subspace $\operatorname{span}(s,v)$ can be written as
\begin{equation}
    \gamma s + \sqrt{1-\gamma^2}\,v,
    \label{eq:target-angular-direction}
\end{equation}
where $\gamma \in [-1,1]$ is the target concept score. This decomposition lets us separate two aspects of steering that are entangled in standard additive interventions: the angular movement toward the concept direction and the change in hidden-state magnitude.

\subsection{Steering direction construction}

For each model and dataset, we construct a concept direction using contrastive mean-difference. We sample $N=256$ positive-negative completion pairs from a held-out direction split and extract residual-stream activations at the last prompt token. The steering direction is the unit-normalized difference between the mean positive activation and the mean negative activation:
\begin{equation}
    s = \frac{\mu_{+} - \mu_{-}}{\|\mu_{+} - \mu_{-}\|}.
    \label{eq:mean-difference-direction}
\end{equation}
The same direction $s$ is used for all steering methods within a model-dataset-fold cell, ensuring that comparisons isolate the geometry of the intervention rather than differences in direction estimation.

\subsection{Steering methods}

We compare six steering operations that differ in whether they preserve the original norm and whether they target the concept score independently for each token. Table~\ref{tab:steering-methods} summarizes the geometric constraints imposed by each method. Below, we describe each of the methods in detail.
% Here, ``fixed concept score'' means that the method enforces the prescribed target $\gamma$ independently for each token.

\begin{table}[t]
\centering
\begin{tabular}{lcc}
\hline
Method & Norm preserved & Tokenwise $\gamma$ \\
\hline
CAA   & $-$ & $-$ \\
CAA-r & $+$ & $-$ \\
CAA-m & $-$ & $+$ \\
S     & $+$ & $+$ \\
AS    & $+$ & $-$ \\
SN    & $-$ & $+$ \\
\hline
\end{tabular}
\caption{
Summary of steering methods by whether they preserve the original hidden-state norm and whether they enforce a fixed per-token concept score.
}
\label{tab:steering-methods}
\end{table}

\smallskip
\noindent \textbf{Concept Activation Addition (CAA).}
The standard additive baseline applies a fixed global perturbation:
\begin{equation}
    y = x + \alpha s.
    \label{eq:caa}
\end{equation}
$\alpha$ is usually treated as a hyperparameter. CAA is neither norm-preserving nor per-token calibrated: it applies the same fixed addition during all generation steps. The achieved concept score varies across tokens depending on the initial norm and alignment of $x$.

\smallskip
\noindent \textbf{Renormalized CAA (CAA-r).}
CAA-r applies the same fixed additive update and then projects the result back to the original norm:
\begin{equation}
    y = r\,\frac{x+\alpha s}{\|x+\alpha s\|}.
    \label{eq:caa-r}
\end{equation}
This isolates the effect of post-hoc norm preservation while retaining the fixed-strength nature of CAA. CAA-r preserves $\|y\|=\|x\|$, but it does not enforce a target concept score for each token.

\smallskip
\noindent \textbf{Matched CAA without renormalization (CAA-m).}
CAA-m chooses a token-specific additive coefficient $\alpha$ so that the normalized output reaches a desired concept score $\gamma$:
\begin{equation}
    y = x + \alpha s,
    \qquad
    \left\langle \frac{y}{\|y\|}, s \right\rangle = \gamma.
    \label{eq:caa-m}
\end{equation}
We compute $\alpha$ using the formula derived in Appendix~\ref{app:caam}. Unlike CAA-r, CAA-m does not renormalize the output. Thus, it exactly controls the angular concept score while allowing the norm to change.

CAA-m and Spherical Steering can be compared in a shared geometric subspace because both operate inside $\operatorname{span}(s,v)$. Once CAA-m chooses $\alpha$ such that the normalized output has concept score $\gamma$, its direction lies on the same ray as the spherical target
\begin{equation}
    \gamma s + \sqrt{1-\gamma^2}\,v.
    \label{eq:caa-m-spherical-ray}
\end{equation}
Therefore, CAA-m and Spherical Steering have the same angular component and differ only in their radial component. If the matched CAA output is additionally renormalized to the original norm, the resulting method, CAA-mr, is exactly equivalent to Spherical Steering. We therefore do not treat CAA-mr as a separate method.

\smallskip
\noindent \textbf{Spherical Steering (S).}
Spherical Steering directly constructs the minimum-geodesic-distance unit direction with target score $\gamma$, then restores the original norm:
\begin{equation}
    y = r\left(\gamma s + \sqrt{1-\gamma^2}\,v\right).
    \label{eq:spherical-steering}
\end{equation}
This method preserves $\|y\|=\|x\|$ exactly and enforces $\langle y/\|y\|, s\rangle=\gamma$ independently for every token.

\smallskip
\noindent \textbf{Additive Spherical (AS).}
Additive Spherical applies a fixed spherical displacement toward the concept direction. Let
\begin{equation}
    \theta = \arccos(c),
    \qquad
    \theta' = \max(\theta-\Delta\theta,0).
    \label{eq:additive-spherical-angle}
\end{equation}
The steered state is
\begin{equation}
    y = r\left(\cos\theta' \, s + \sin\theta' \, v\right).
    \label{eq:additive-spherical}
\end{equation}
AS preserves the norm and the residual direction, but it does not target the same final concept score for every token. Instead, the resulting score depends on the token's initial angle to $s$.

\smallskip
\noindent \textbf{Spherical Steering with Norm Scaling (SN).}
Finally, we introduce an explicit radial parameter $\beta$ on top of Spherical Steering:
\begin{equation}
    y = \beta r\left(\gamma s + \sqrt{1-\gamma^2}\,v\right).
    \label{eq:spherical-norm-scaling}
\end{equation}
When $\beta=1$, SN reduces exactly to S. For $\beta\neq1$, the angular component is unchanged while the norm is scaled by a fixed multiplicative factor. This lets us test whether the norm acts primarily as a stability parameter once semantic angular control is fixed.

Our experimental design isolates the roles of angular control and norm modification through four controlled experiments.

\paragraph{1. Hidden-state norm variation.}
First, we measure hidden-state norm variation across layers and token populations. For each model, we sample examples from multiple corpora and compute the coefficient of variation of $\|x\|$ for the last prompt token, all prompt tokens, and generated tokens. This experiment characterizes the radial geometry of the representation space and determines whether norm preservation is a meaningful constraint.

\paragraph{2. Angular versus radial concept encoding.}
Second, we test whether concept information is encoded primarily in direction or magnitude. We train three linear probes: one on raw hidden states $h$, one on normalized hidden states $h/\|h\|$, and one on the scalar norm $\|h\|$. If normalized probes match raw probes while norm-only probes remain near chance, this indicates that concept information is primarily angular rather than radial.

\paragraph{3. Steering at matched angular control.}
Third, we compare steering methods under matched angular control. For per-token methods, we set a target concept score $\gamma$. For fixed-strength methods, we calibrate the global strength parameter by binary search so that the mean achieved concept score on evaluation activations matches the desired target $\bar{\gamma}$. We then compare downstream task performance, per-token concept-score variance, norm ratio $\|y\|/\|x\|$, perplexity, and general capability metrics. This experiment distinguishes three possible explanations for steering behavior: per-token precision, angular displacement, and norm preservation.

\paragraph{4. Isolating the role of norm scaling.}
Fourth, we isolate the role of the norm using SN. Holding $\gamma$ fixed, we vary only the multiplicative norm scale $\beta$. Because $\beta$ changes only the radius and leaves the angular concept score fixed, this experiment directly tests whether modest norm changes improve generation stability without changing semantic control.

All steering directions are computed on held-out direction splits and evaluated on separate held-out examples. Where methods require calibration, we perform binary search over the steering parameter before measuring downstream behavior. For CAA and CAA-r, the searched parameter is $\alpha$; for AS, it is $\Delta\theta$; for CAA-m, it is the token-specific $\alpha$ needed to achieve $\gamma$.

Together, these experiments provide a controlled comparison between interventions that alter direction, norm, or both. By matching either per-token concept score or mean concept score across methods, we can determine whether steering success is explained by angular movement alone, strict norm preservation, or a two-parameter interaction between angle and radius.

\section{Experiments}
\label{sec:experiments}

\subsection{Evaluation setup}
\label{subsec:evaluation_setup}

\noindent \textbf{Models and steering layer.}
We evaluate all methods on seven transformer language models spanning 1B to 70B parameters: Llama-3.1-8B-Instruct, Qwen2.5-7B-Instruct, Gemma-2-9B-it, Llama-3.1-8B, Llama-3.2-1B-Instruct, Qwen2.5-3B-Instruct, and Llama-3.1-70B-Instruct. For each model, steering is applied to the residual-stream output at 75\% depth. This gives steering layers 24, 21, 31, 24, 12, 27, and 60 respectively. We use a single forward hook at this layer, replacing each hidden state $x$ with a steered state $y$ at every token position during generation.

\smallskip
\noindent \textbf{Datasets and task metrics.}
We evaluate steering on four concept datasets: TruthfulQA for truthfulness, SST-2 for sentiment, CivilComments for toxicity, and IMDB for sentiment. For TruthfulQA, we use closed-form multiple-choice metrics, primarily MC1. For SST-2 and IMDB, we measure the positive rate of generated continuations. For CivilComments, we measure non-toxicity using a toxicity classifier. For generation-based evaluations, we sample 128 tokens using nucleus sampling with $p=0.95$ and temperature $T=0.7$. Dataset and benchmark details are provided in \appref{app:datasets}.

\smallskip
\noindent \textbf{Quality and capability metrics.}
To measure whether steering damages general language-model behavior, we compute perplexity on 200 held-out WikiText-103 passages with maximum length 512. We report perplexity as a ratio relative to the unsteered baseline for the same model, dataset, and fold. We also evaluate MMLU accuracy using log-probability ranking on a fixed subset of 300 items, providing an auxiliary measure of retained model capability.

\smallskip
\noindent \textbf{Calibration protocol.}
For per-token methods, we sweep target concept scores
$\gamma \in \{0.1, 0.3, 0.5, 0.7\}$.
For fixed-strength methods, we calibrate the global steering parameter so that the mean achieved
concept score on evaluation activations matches the desired target $\bar{\gamma}$. Specifically, we
binary-search $\alpha$ for CAA and CAA-r, and $\Delta\theta$ for AS. For SN, we hold the angular
target fixed and sweep
% \vspace{-2.0cm}
$\beta \in \{0.9, 1.0, 1.1, 1.2\}.$
% \vspace{-2.5cm}
All reported comparisons use held-out direction splits and held-out evaluation examples. Unless
otherwise stated, results are aggregated over seven models, four datasets, one seed, and two folds.

\subsection{Experimental results}
\label{subsec:experimental_results}

\begin{figure}[h]
\centering
\includegraphics[width=0.48\linewidth]{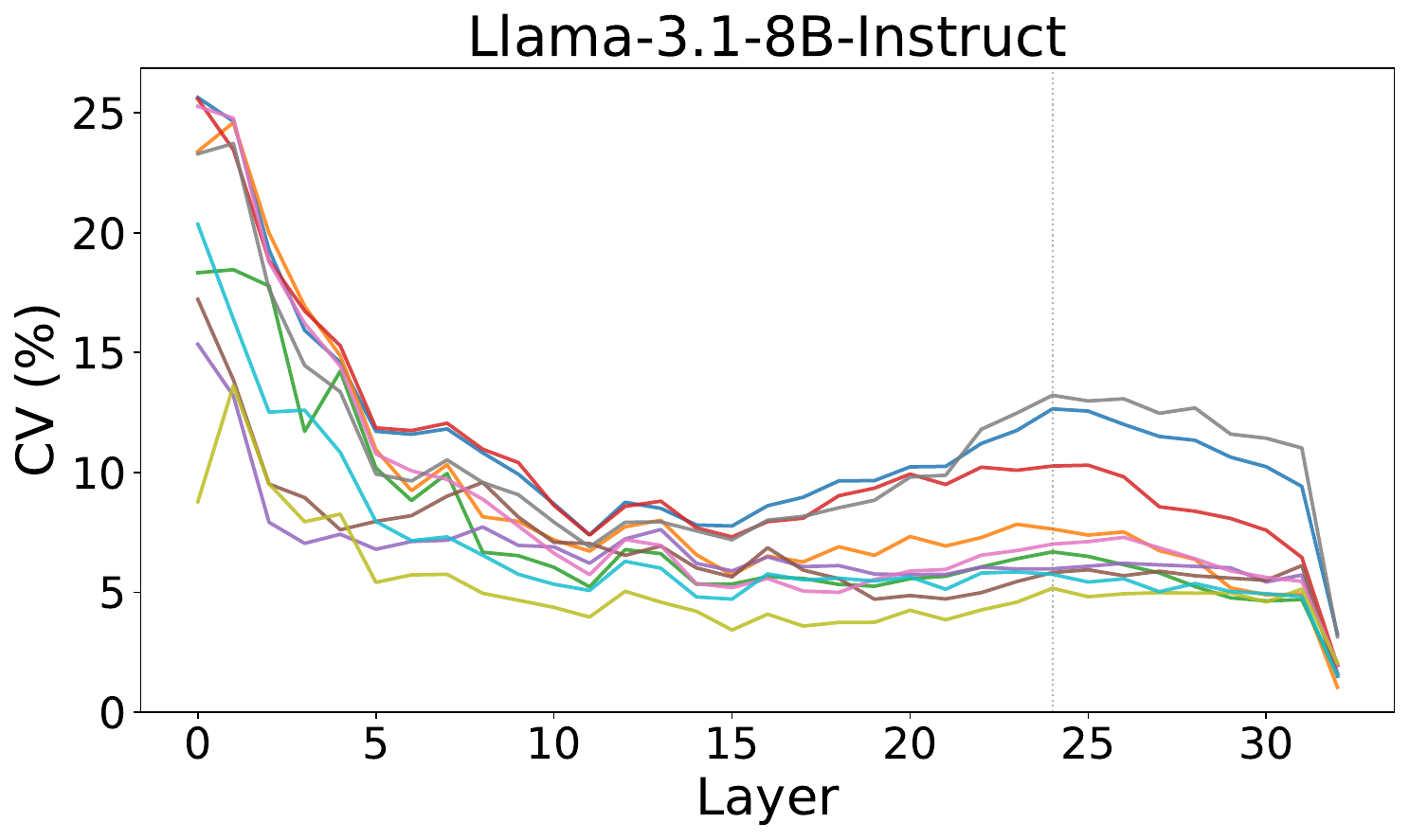}
\hfill
\includegraphics[width=0.48\linewidth]{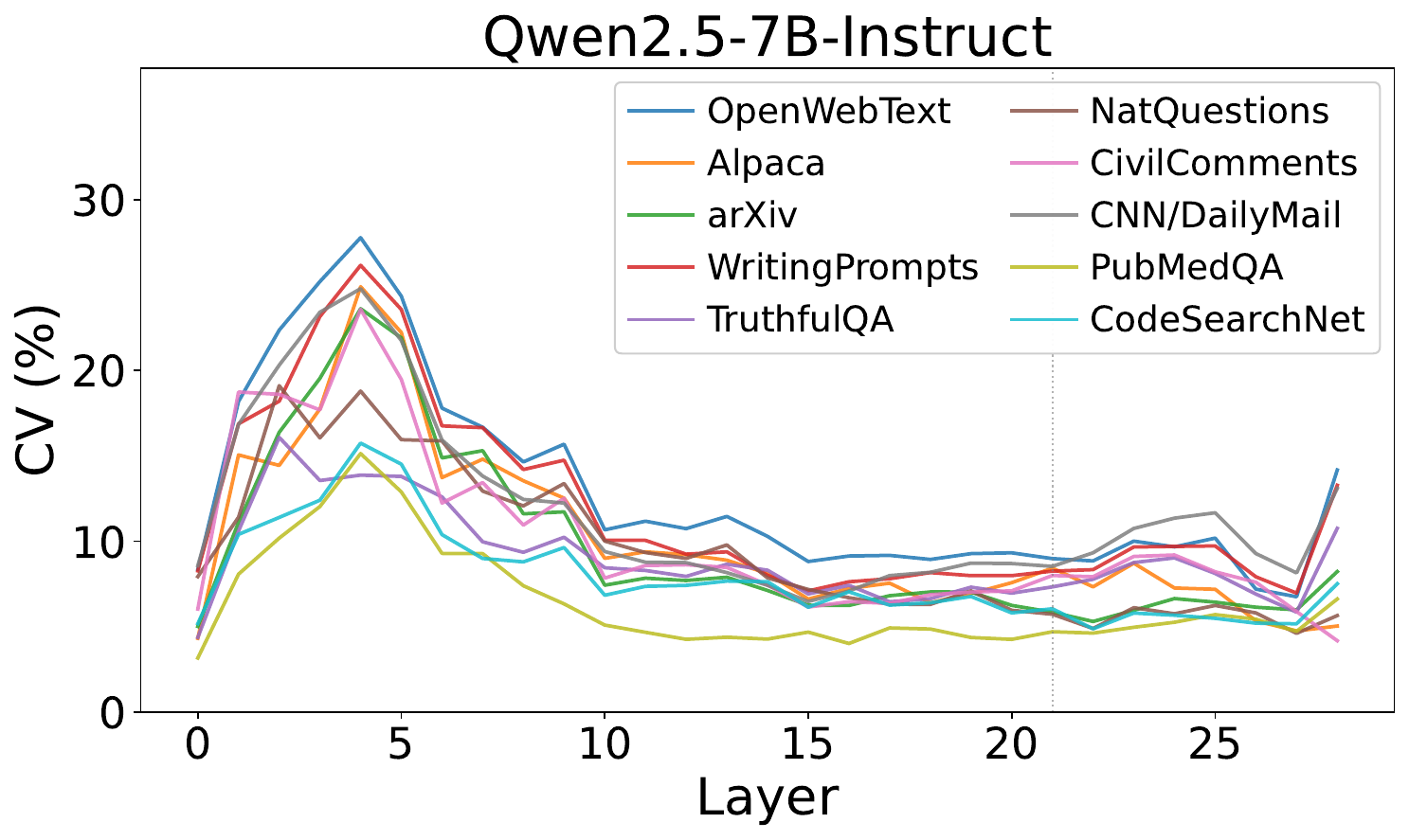}
\\[4pt]
\includegraphics[width=0.48\linewidth]{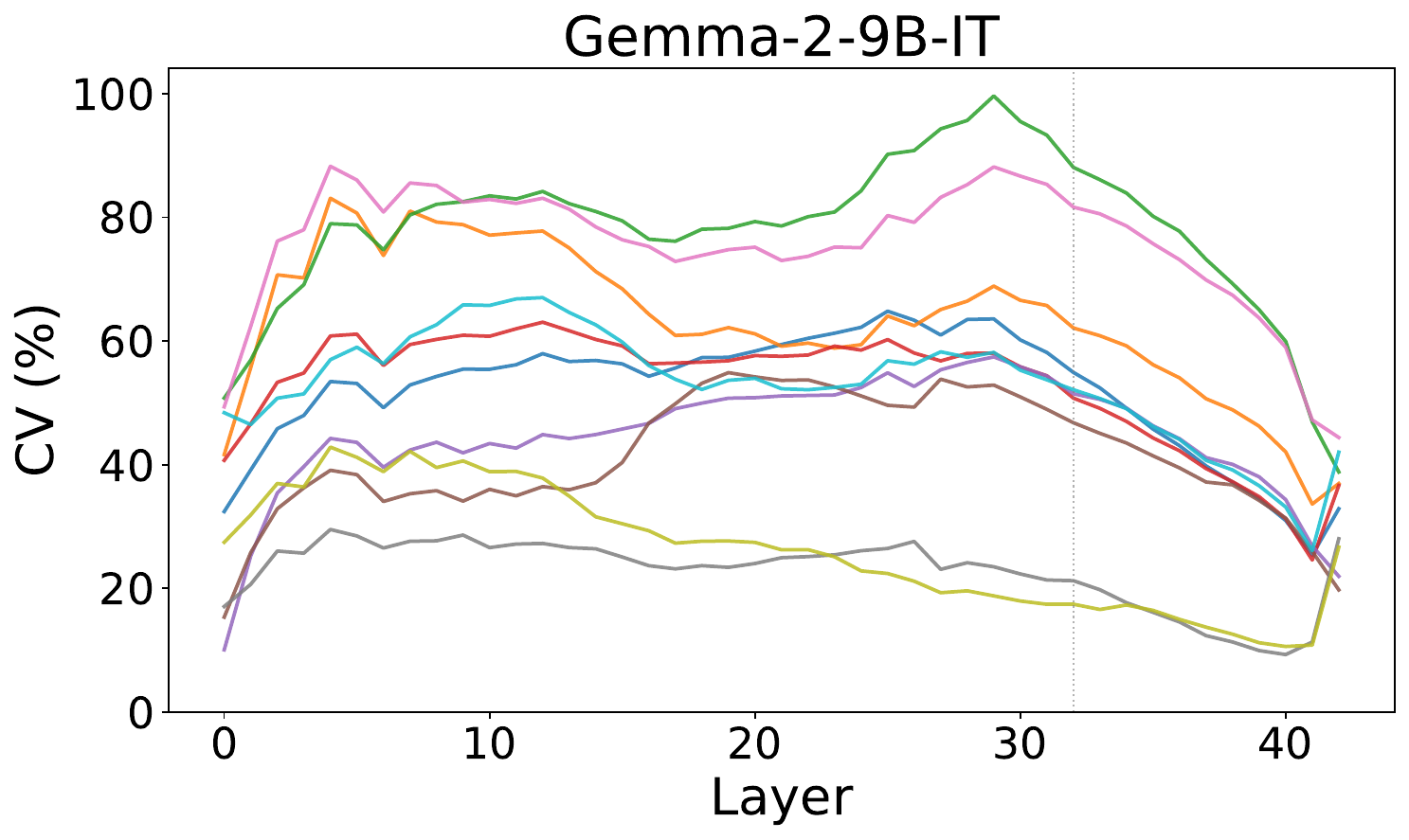}
\hfill
\includegraphics[width=0.48\linewidth]{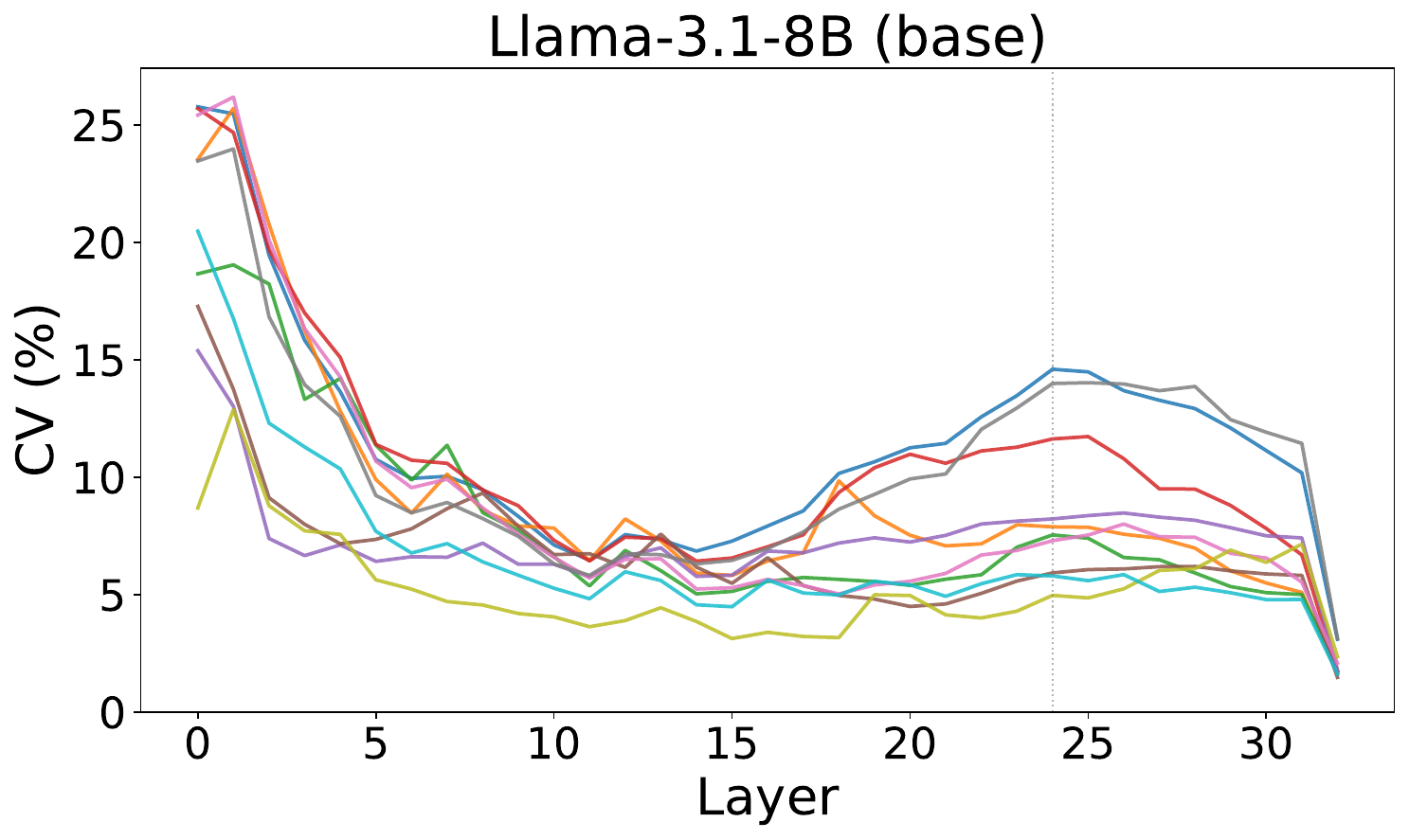}
\\[4pt]
\includegraphics[width=0.48\linewidth]{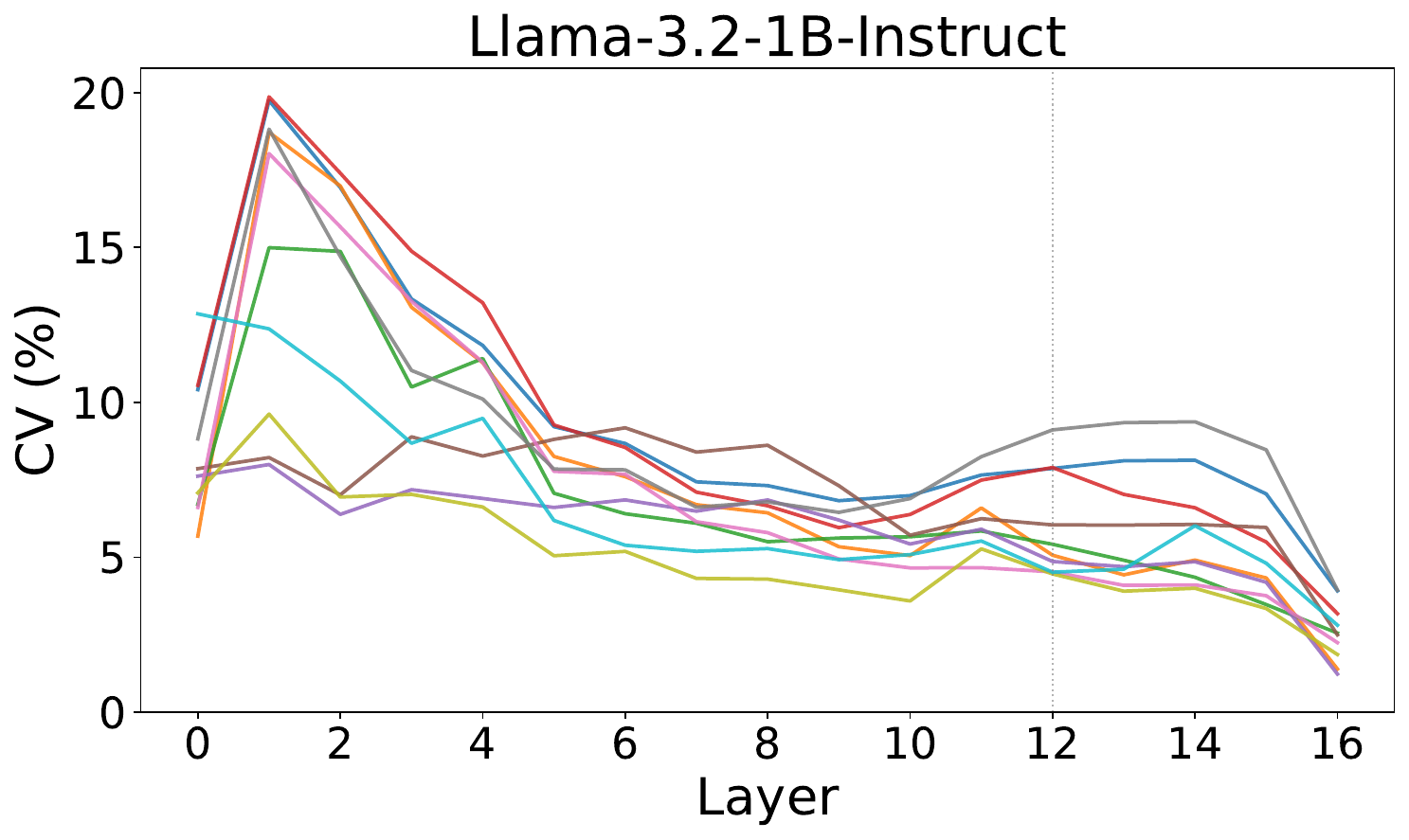}
\hfill
\includegraphics[width=0.48\linewidth]{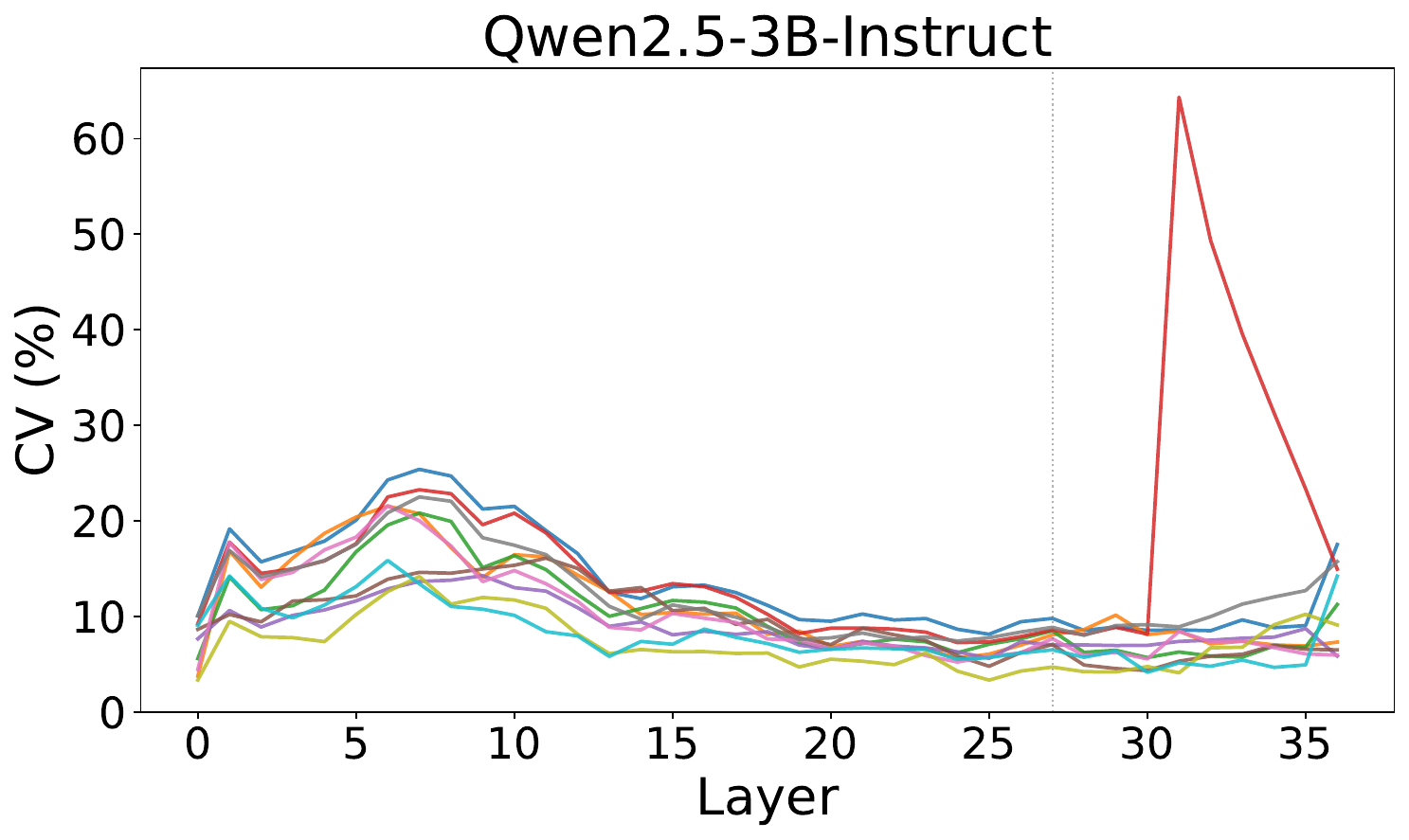}
\\[4pt]
\includegraphics[width=0.48\linewidth]{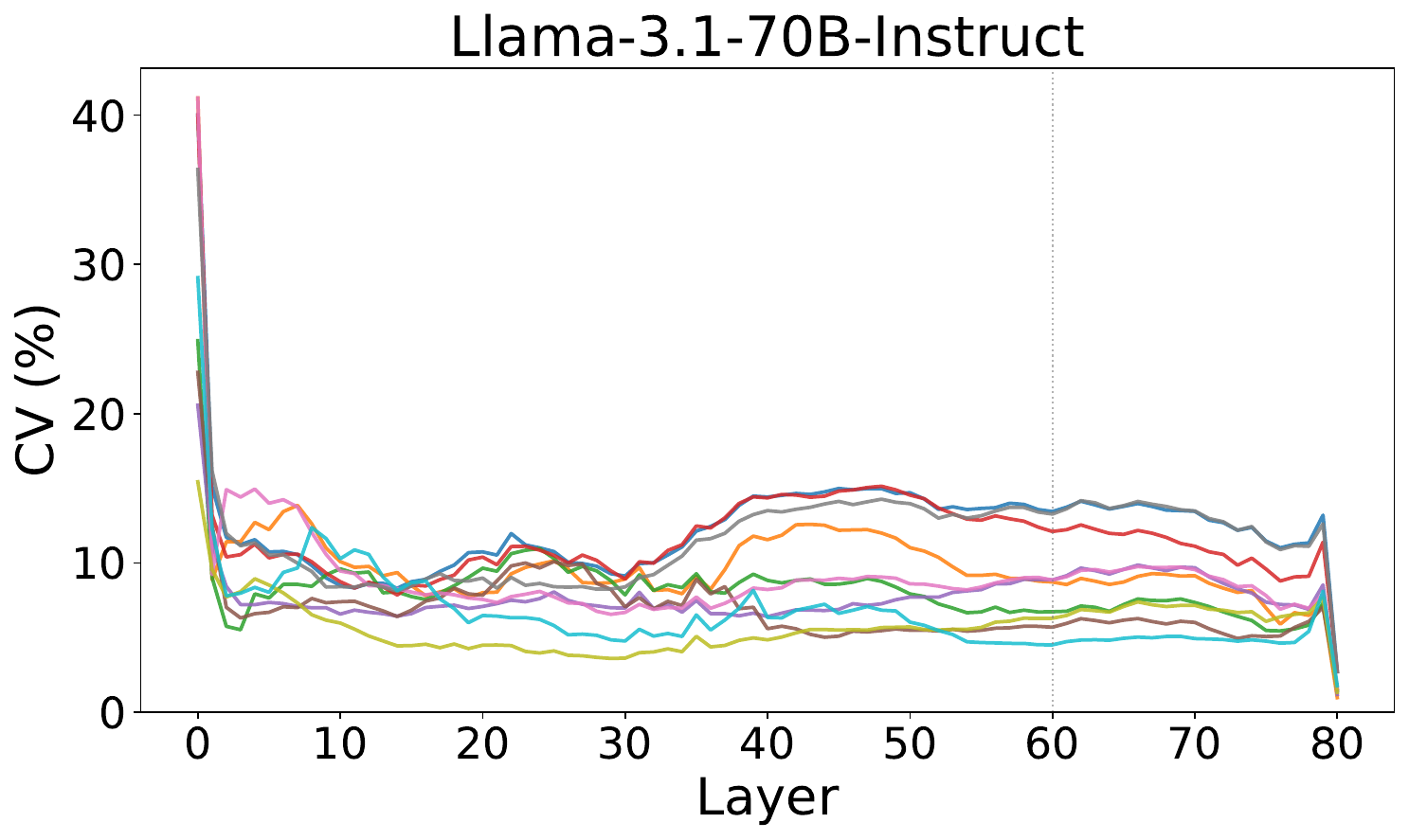}
\hfill
\includegraphics[width=0.48\linewidth]{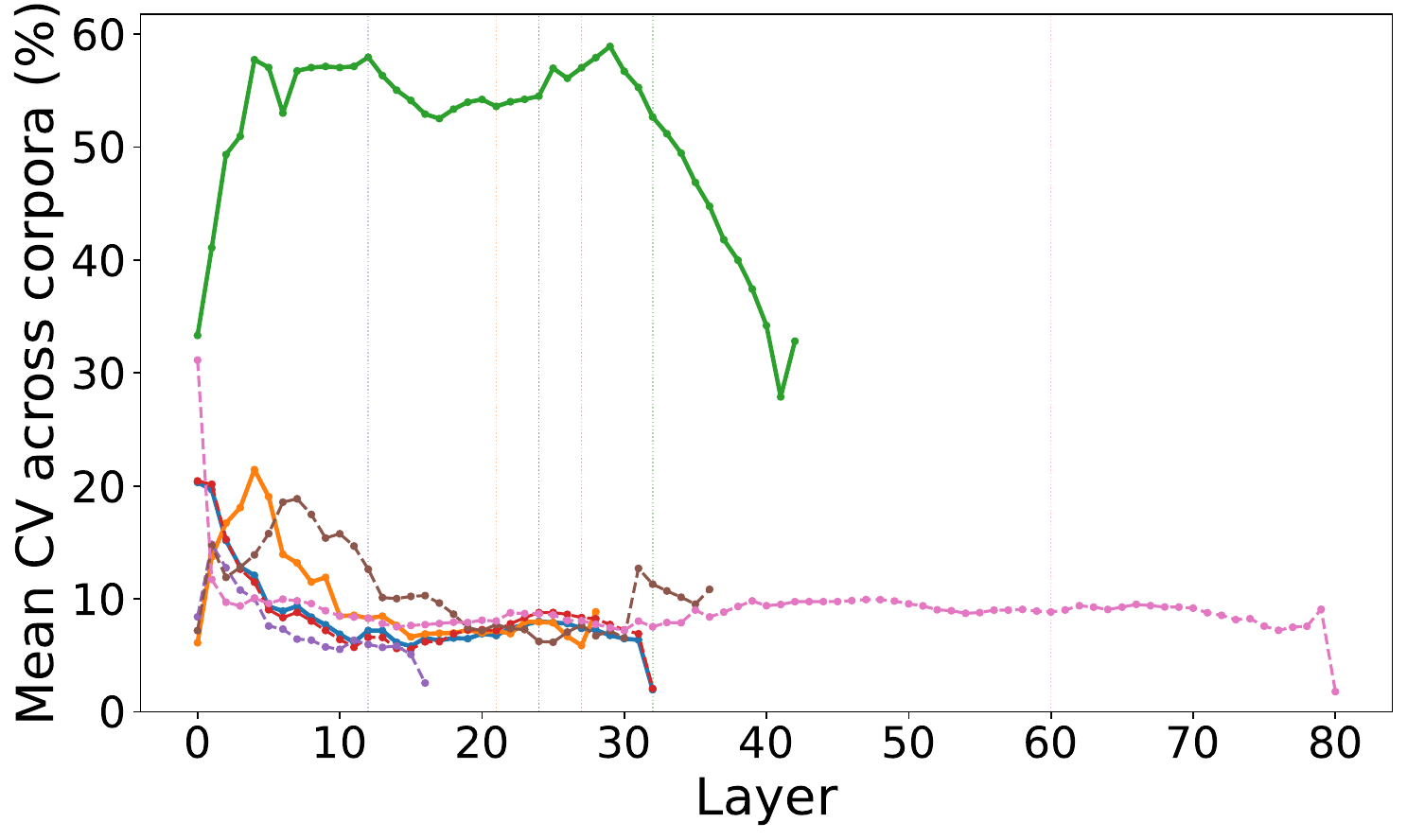}
\caption{T1: CV of hidden-state norms vs.\ layer for all 7 models, 10 corpora. Grey dotted = $L_{75}$ steering layer. Bottom right: combined mean CV across corpora.}
\label{fig:T1_cv}
\end{figure}

\paragraph{Hidden-state norms vary across layers and architectures.}
We first examine whether activation norms can be treated as approximately constant during
steering. Figure~\ref{fig:T1_cv} reports the coefficient of variation of last-prompt-token
hidden-state norms across layers, models, and corpora. The results show that norm concentration is
architecture-dependent: Llama and Qwen models generally have relatively concentrated norms at
middle and later layers, while Gemma exhibits much larger norm variation across most layers. We hypothesize that this difference is largely due to Gemma's post-norm architecture. Across models, the activations after the last transformer block
consistently have the lowest coefficient of variation, indicating that norm concentration increases
toward the final layers. This
indicates that the radial component is not a universally negligible part of the representation space. Additional layer-wise norm statistics are reported in \appref{app:norm_variation}.

Importantly, norm variation by itself does not determine whether a concept is encoded in the norm
or in the direction. Rather, this experiment motivates treating the norm as a separate geometric
degree of freedom: even when semantic information is primarily angular, preserving or modifying
the radius may still affect generation stability.

\begin{figure}[t]
\centering
\includegraphics[width=\linewidth]{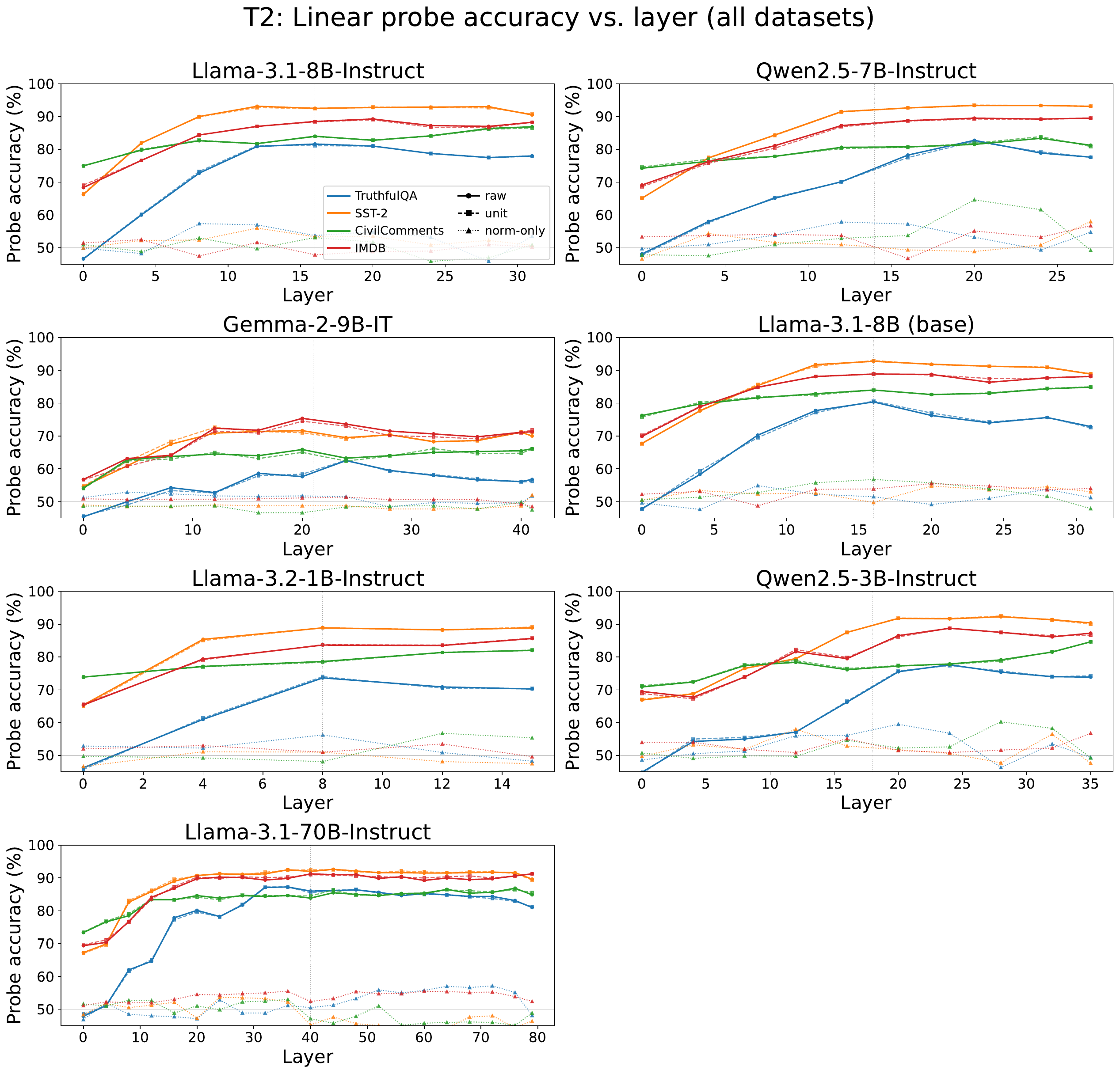}
\caption{
Linear probe accuracy versus layer for all four concept datasets. Each dataset contains three
probe variants: raw hidden states $h$, normalized hidden states $h/\|h\|$, and norm-only features
$\|h\|$. Raw and normalized curves nearly overlap, while norm-only probes remain close to chance,
indicating that the evaluated concepts are encoded primarily in direction.
}
\label{fig:directional_encoding}
\end{figure}

\paragraph{Concept information is primarily angular.}
We next test whether concept-discriminative information is encoded in the direction or the
magnitude of hidden states. As shown in Figure~\ref{fig:directional_encoding}, for each model and
dataset, we train linear probes on three representations: raw hidden states $h$, normalized hidden
states $h/\|h\|$, and scalar norms $\|h\|$. Across all models and concept datasets, normalized
probes closely match raw probes, while norm-only probes remain near chance. This supports the
central geometric assumption that concept information is primarily represented in angular
directions rather than in hidden-state magnitudes. Additional directional-encoding results are provided in \appref{app:directional_encoding}.

\paragraph{Matched additive steering and spherical steering share the same angular target but differ in norm.}
We next compare CAA-m and S at matched per-token target $\gamma$. Both methods steer inside
the same two-dimensional subspace $\operatorname{span}(s,v)$ and reach the same normalized
concept direction. Their difference is radial: S restores the original norm, while CAA-m leaves the
additive norm change intact. This comparison therefore isolates the effect of norm change while
holding angular control fixed. Further comparisons are provided in \appref{app:t3a_s_vs_caam}.

\begin{figure}[t]
\centering
\includegraphics[width=\linewidth]{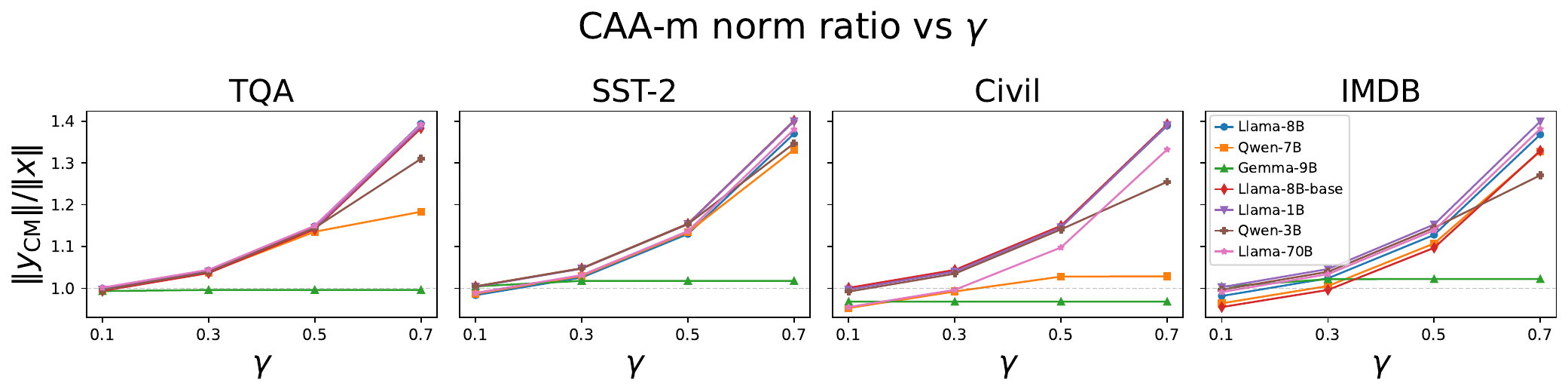}
\caption{
Norm ratio $\|y\|/\|x\|$ for CAA-m at matched per-token target $\gamma$.
}
\label{fig:caam_norm_ratio}
\end{figure}

Figure~\ref{fig:caam_norm_ratio} shows that CAA-m produces only mild norm inflation at low and
moderate steering strengths, but the effect grows at high $\gamma$.

Despite matching the same angular target, S and CAA-m differ strongly in generation stability.
At high $\gamma$, strict norm preservation can produce large perplexity penalties and substantial
capability loss, whereas CAA-m often retains lower perplexity and higher MMLU accuracy. This
shows that preserving the original norm is not always the most stable choice once the angular edit
becomes large.

\begin{figure}[t]
\centering
\includegraphics[width=\linewidth]{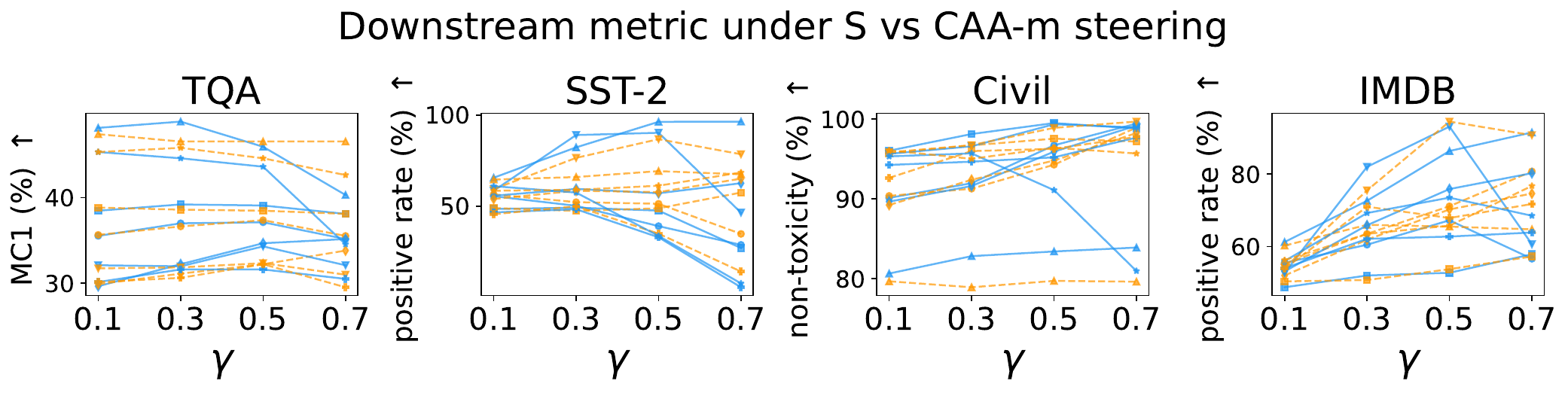}\\[4pt]
\includegraphics[width=\linewidth]{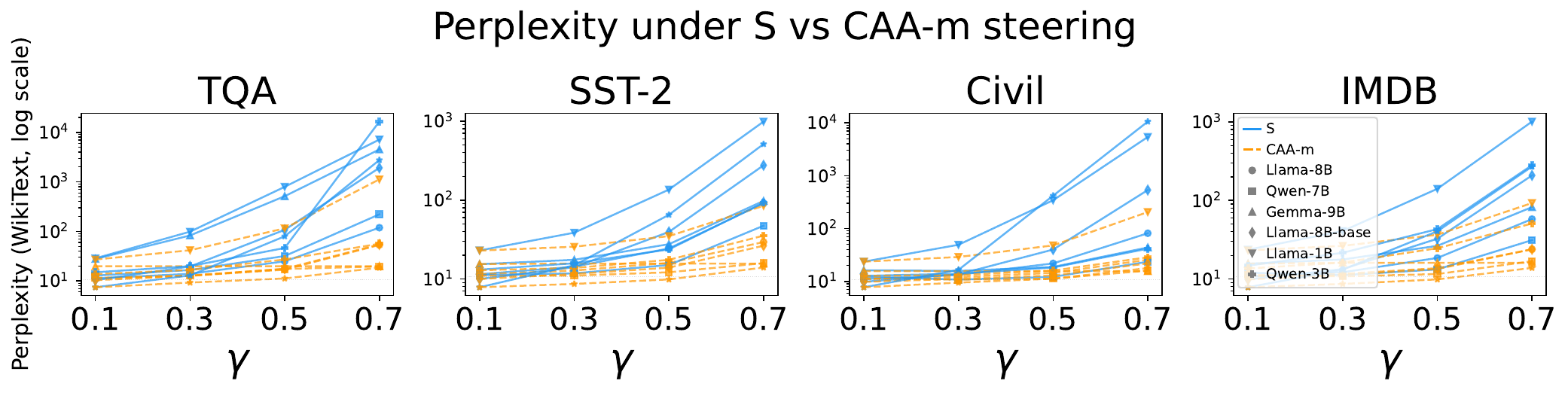}\\[4pt]
\includegraphics[width=\linewidth]{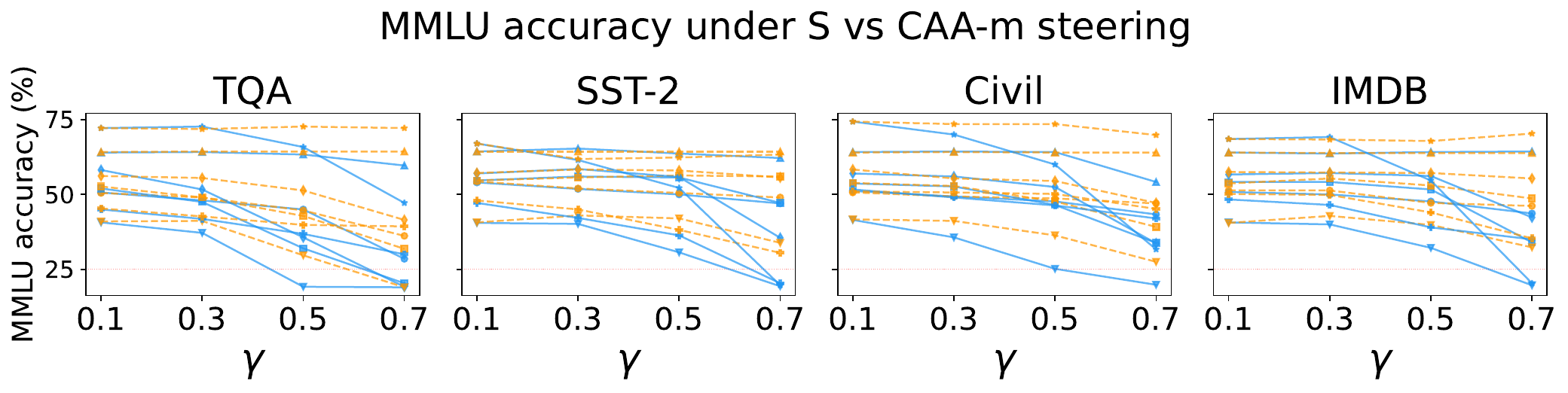}
\caption{
Downstream task metric, WikiText-103 perplexity, and MMLU accuracy under S and CAA-m at
matched per-token $\gamma$. The two methods implement nearly identical angular control, but
they differ in radial behavior. At high steering strengths, S incurs much larger perplexity penalties,
while CAA-m better preserves generation stability and general capability.
}
\label{fig:s_vs_caam}
\end{figure}

\paragraph{Norm preservation alone does not explain stability in fixed-strength steering.}
We next isolate the fixed-strength family: CAA, CAA-r, and AS. Unlike S and CAA-m, these
methods do not enforce the same concept score independently for each token. Instead, each method
uses a single global steering parameter, calibrated so that the mean achieved concept score matches
the target level. This comparison tests whether preserving the hidden-state norm is sufficient to
explain downstream stability. Additional results are provided in \appref{app:t3b_fixed_strength}.

The first comparison is between CAA and CAA-r. These methods have the same normalized output
direction after the additive update; CAA-r only rescales the resulting vector back to the original
norm. As a result, their downstream behavior is very similar across steering strengths. This shows
that post-hoc renormalization is not, by itself, a reliable source of improved stability.

The second comparison is between CAA-r and AS. Both methods preserve the hidden-state norm,
but they produce different token-level angular profiles. CAA-r applies a fixed additive perturbation
before renormalization, so the resulting angular displacement depends on the token's initial norm
and alignment with the steering direction. AS applies a fixed spherical displacement, so its effect is
distributed differently across tokens. The fact that these two norm-preserving methods behave
differently shows that norm preservation alone cannot explain the steering quality trade-off.
Instead, the per-token distribution of achieved concept scores is an important part of the geometry.

\begin{figure*}[th!]
\centering
\includegraphics[width=\linewidth]{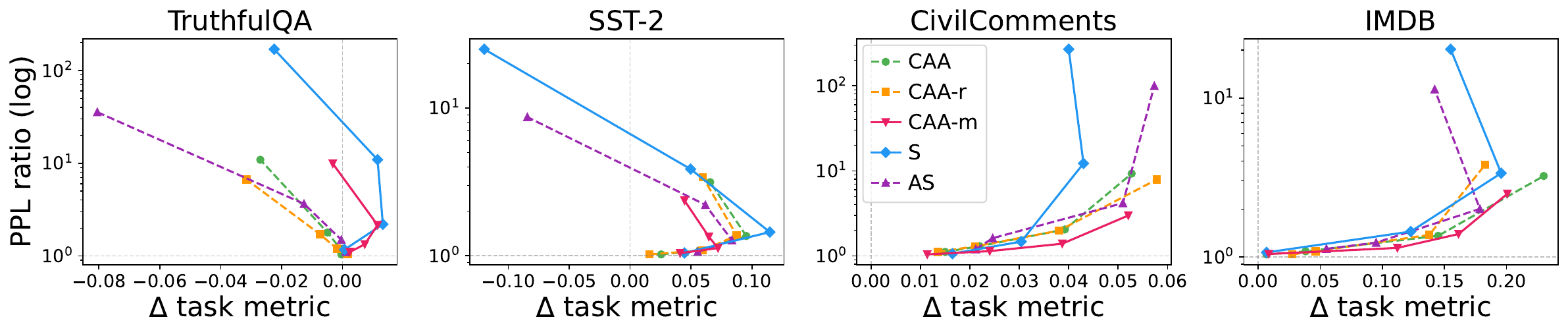}
\caption{
Per-dataset Pareto curves for all methods. The same qualitative pattern appears across datasets:
CAA-m provides a strong high-control, low-perplexity trade-off, while S suffers a large perplexity
increase at high steering strengths.
}
\label{fig:pareto_per_dataset}
\end{figure*}

\paragraph{The Pareto frontier depends on both angular precision and radial behavior.}
We then compare all five main methods: CAA, CAA-r, CAA-m, S and AS. Fixed-strength methods
are calibrated to matched mean concept score, while per-token methods directly enforce the target
score for each token. We plot downstream task improvement against WikiText-103 perplexity ratio,
so better methods move toward higher task improvement and lower perplexity.
Figure~\ref{fig:pareto_per_dataset} shows the Pareto comparison separately for each dataset.

As shown in Figure~\ref{fig:pareto_per_dataset}, these results
suggest that steering should not be reduced to a binary choice between preserving and changing
the norm. CAA-m and S have the same angular target, yet CAA-m is much more stable at high
$\gamma$. Conversely, CAA-r and AS both preserve norm, but AS produces much higher perplexity
at large steering strengths. Particular, S achieves the highest downstream task score even at $\gamma=0.5$, showing that strict per-token angular targeting can be highly effective for semantic control. \textbf{The relevant design space is therefore two-dimensional}: angular control
determines the semantic effect of steering, while norm scale strongly influences whether the model
can continue generating coherently.

\paragraph{Norm scaling acts as a stability lever.}
Finally, we test this interpretation directly by adding an explicit multiplicative norm scale $\beta$ on top of S. This gives SN:
\[
    y = \beta r \left(\gamma s + \sqrt{1-\gamma^2}v\right).
\]
Changing $\beta$ leaves the angular concept score fixed while changing only the radius of the steered representation. Thus, within this controlled intervention family, differences across $\beta$ values reflect the effect of radial scaling under a fixed semantic direction. 

Figure~\ref{fig:beta_effect} shows that $\beta$ has only a small effect on the task metric but a large effect on perplexity at high $\gamma$. Moving from $\beta=1.0$ to $\beta=1.2$ improves perplexity by roughly $1.8\times$ at $\gamma=0.7$, while task metrics remain within about $2.5$ percentage points across the tested $\beta$ values. We hypothesize that this effect arises because the hidden-state norm partly determines the representational capacity available to the model at that token. When steering strongly toward a concept direction while strictly preserving the original norm, a large fraction of the fixed-radius representation may be devoted to expressing the steered concept, leaving less effective capacity for other information needed to maintain fluent and contextually coherent generation. A modest norm increase may compensate for this by allowing the target concept to be expressed without compressing the remaining information into the same radius.

Overall, these experiments support \textbf{a two-parameter view of activation steering}. The angular component controls the intended concept, as shown by the probe results and by the matched behavior of S and CAA-m in concept space. The radial component controls stability: preserving the original norm is sometimes useful, but at high steering strengths a modest norm increase can substantially reduce perplexity without materially changing the semantic steering effect. Additional beta-sweep results are provided in \appref{app:t4_beta_sweep}.

\section{Conclusion}

We presented a geometric account of activation steering that separates two effects entangled by additive interventions: angular movement toward a concept direction and radial change in hidden-state norm. This explains why a single additive coefficient is hard to interpret: the same coefficient can induce different angular shifts and norm changes depending on each token's initial geometry.

Across seven language models and four concept datasets, we find that the evaluated concepts are represented primarily in activation direction. Normalized probes closely match raw probes, while norm-only probes remain near chance, supporting the view that semantic control is largely angular.

At the same time, norm preservation is not always the right constraint. Even with fixed angular concept score, radial changes can strongly affect perplexity and capability preservation. Strict norm preservation can become unstable at high steering strengths, while modest norm increases reduce degradation without materially changing the semantic effect.

Overall, our findings reframe activation steering as a two-parameter intervention over angle and radius. Angle controls the intended concept, while radius controls intervention stability. This explains why methods with similar concept-level effects can behave differently and suggests a more interpretable basis for future token-wise steering methods. Effective steering requires choosing not only where to point a representation, but also how much representational scale to give it.

\section{Limitations}
\label{sec:limitations}

Our study has several limitations. First, we apply steering at a single fixed layer, chosen at 75\%
depth for each model. Although this gives a controlled comparison across methods, the optimal
angle--norm trade-off may vary across layers.

Second, our experiments cover a limited set of models and concepts. We evaluate Llama, Qwen,
and Gemma models on truthfulness, sentiment, and toxicity-related steering, but other architectures
or more complex behaviors may exhibit different geometry.

Third, all methods use the same contrastive mean-difference steering direction. This isolates the
effect of the intervention geometry, but does not test whether the conclusions hold for other ways
of estimating steering directions.

Finally, our norm-scaling experiments use a small discrete set of $\beta$ values. The results show
that the norm is an important stability parameter, but they do not provide an automatic rule for
choosing the best norm scale for a new model, layer, or task.

% Bibliography entries for the entire Anthology, followed by custom entries
%\bibliography{custom,anthology-overleaf-1,anthology-overleaf-2}

% Custom bibliography entries only
\bibliography{custom}
\clearpage
\appendix
\tableofcontents

\section*{Acknowledgments}
This work was supported by a Google DeepMind PhD Studentship, and the work utilized Queen Mary’s Andrena HPC facility, supported by QMUL Research-IT. This work was also supported by the Engineering and Physical Sciences Research Council [grant number EP/Y009800/1], through funding from Responsible Ai UK (KP0016). 

\section{Additional Norm-Variation Analysis}
\label{app:norm_variation}

The main text reports the layerwise pattern of last-prompt-token norm variation. This appendix
provides the supporting details. We first report per-corpus CV at the 75\%-depth layer, and then
expand the analysis to prompt and generation positions. These per-position plots separate
cross-sample norm variation from position-dependent norm effects, which can be hidden by
aggregate statistics.

\paragraph{Per-corpus norm variation.}
Table~\ref{tab:app_t1_per_corpus} reports the per-corpus CV of last-prompt-token hidden-state
norms at the 75\%-depth layer. The same qualitative pattern as in the main text holds across
corpora: Llama and Qwen models usually have moderate CV, while Gemma has substantially larger
variation because of its post-norm architecture.

\begin{table*}[t]
\centering
\scriptsize
\caption{
Per-corpus CV of last-prompt-token hidden-state norms at the 75\%-depth layer.
}
\label{tab:app_t1_per_corpus}
\resizebox{\textwidth}{!}{
\begin{tabular}{lccccccc}
\toprule
Corpus & Llama-8B & Qwen-7B & Gemma-9B & Llama-8B (base) & Llama-1B & Qwen-3B & Llama-70B \\
\midrule
OpenWebText        & 12.65\% & 8.99\% & 54.93\% & 14.60\% & 7.86\% & 9.80\% & 13.44\% \\
Alpaca             &  7.63\% & 8.43\% & 62.12\% &  7.89\% & 5.06\% & 8.08\% &  8.72\% \\
arXiv              &  6.68\% & 5.85\% & 88.08\% &  7.54\% & 5.42\% & 8.50\% &  6.72\% \\
WritingPrompts     & 10.26\% & 8.25\% & 50.77\% & 11.63\% & 7.89\% & 8.58\% & 12.10\% \\
TruthfulQA         &  5.98\% & 7.33\% & 51.48\% &  8.23\% & 4.86\% & 7.03\% &  8.85\% \\
Natural Questions  &  5.82\% & 5.73\% & 46.79\% &  5.93\% & 6.04\% & 7.06\% &  5.69\% \\
CivilComments      &  7.01\% & 8.00\% & 81.66\% &  7.30\% & 4.53\% & 7.68\% &  8.85\% \\
CNN/DailyMail      & 13.21\% & 8.53\% & 21.25\% & 14.00\% & 9.11\% & 8.87\% & 13.27\% \\
PubMedQA           &  5.16\% & 4.70\% & 17.45\% &  4.97\% & 4.46\% & 4.71\% &  6.29\% \\
CodeSearchNet      &  5.75\% & 6.05\% & 52.12\% &  5.80\% & 4.51\% & 6.51\% &  4.51\% \\
\bottomrule
\end{tabular}
}
\end{table*}

\begin{figure*}[t]
\centering
\includegraphics[width=\textwidth]{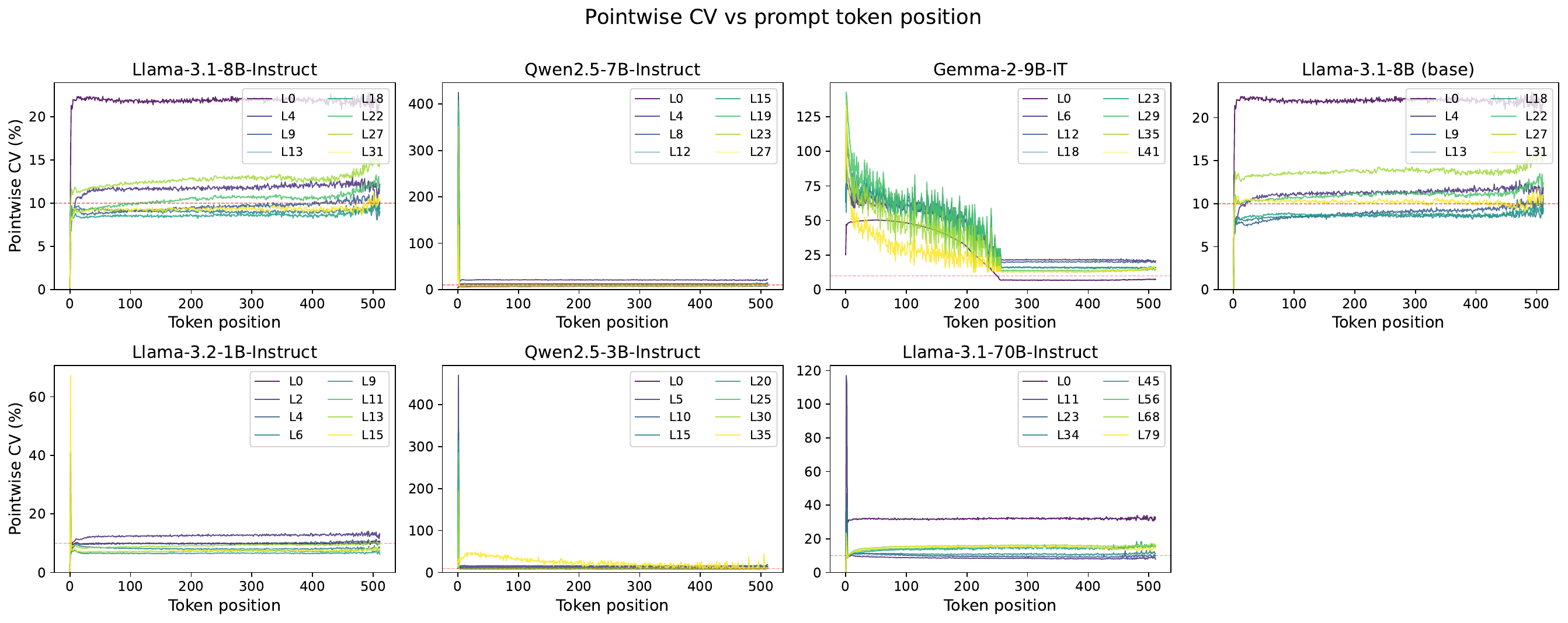}
\caption{
Pointwise CV of hidden-state norms across prompt-token positions. The first prompt positions,
especially position~0, show strong architecture-dependent effects; later positions settle to a more
stable plateau.
}
\label{fig:app_t1_pointwise_prompt}
\end{figure*}

\paragraph{Prompt-token positions.}
Figure~\ref{fig:app_t1_pointwise_prompt} shows pointwise CV across prompt positions. The largest
position-specific effect appears at the beginning of the prompt: in Llama and Qwen models, the
first token behaves like an attention-sink position and has a distinct norm distribution. After the
first few tokens, CV settles to a more stable plateau. Gemma remains different, with elevated
variation across many layers because of its post-norm architecture.

\begin{figure*}[t]
\centering
\includegraphics[width=\textwidth]{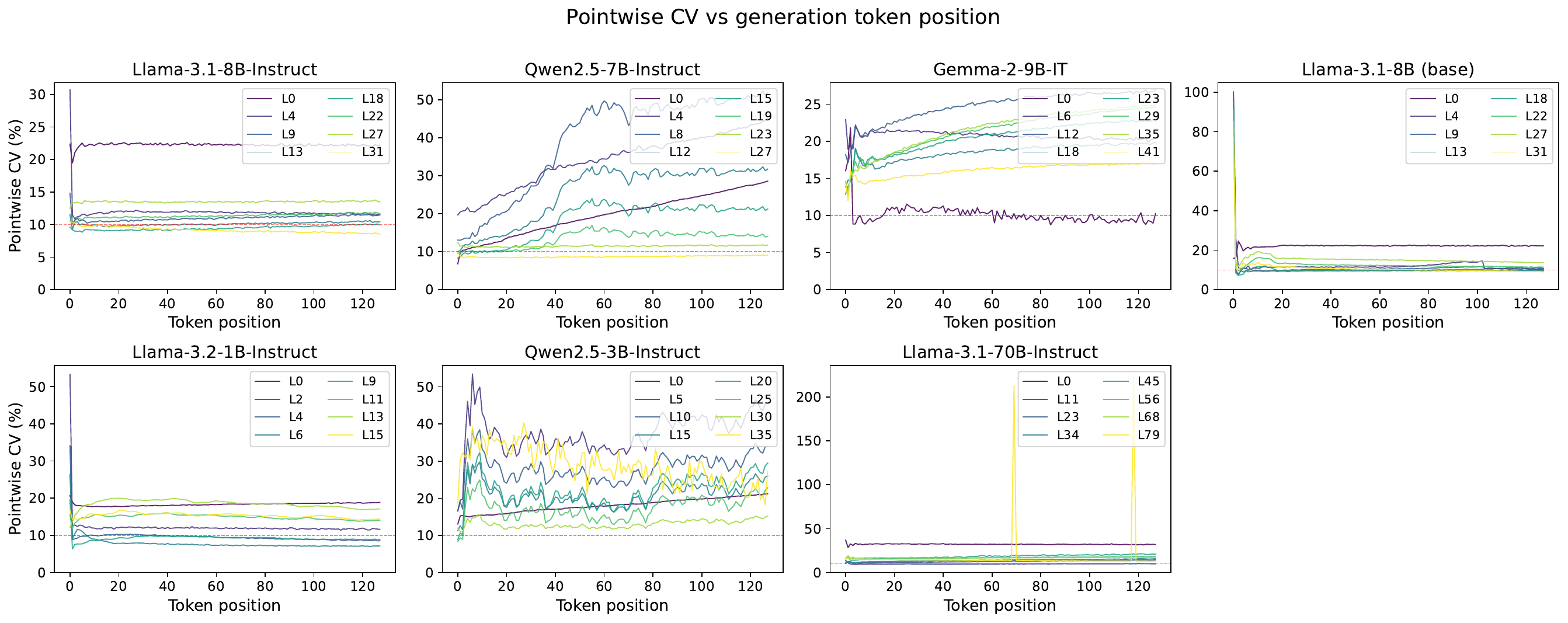}
\caption{
Pointwise CV of hidden-state norms across generation-token positions. Instruction-tuned models
show relatively stable generation-token CV, while the Llama base model has elevated variation at
later layers.
}
\label{fig:app_t1_pointwise_gen}
\end{figure*}

\paragraph{Generation-token positions.}
Figure~\ref{fig:app_t1_pointwise_gen} shows the same analysis for generated tokens. Compared
with prompt tokens, generation positions are more stable for most instruction-tuned models, which
is the relevant regime for the steering hook during decoding. The Llama base model is less stable
under unconstrained generation and shows larger CV at later layers.

\begin{figure*}[t]
\centering
\includegraphics[width=\textwidth]{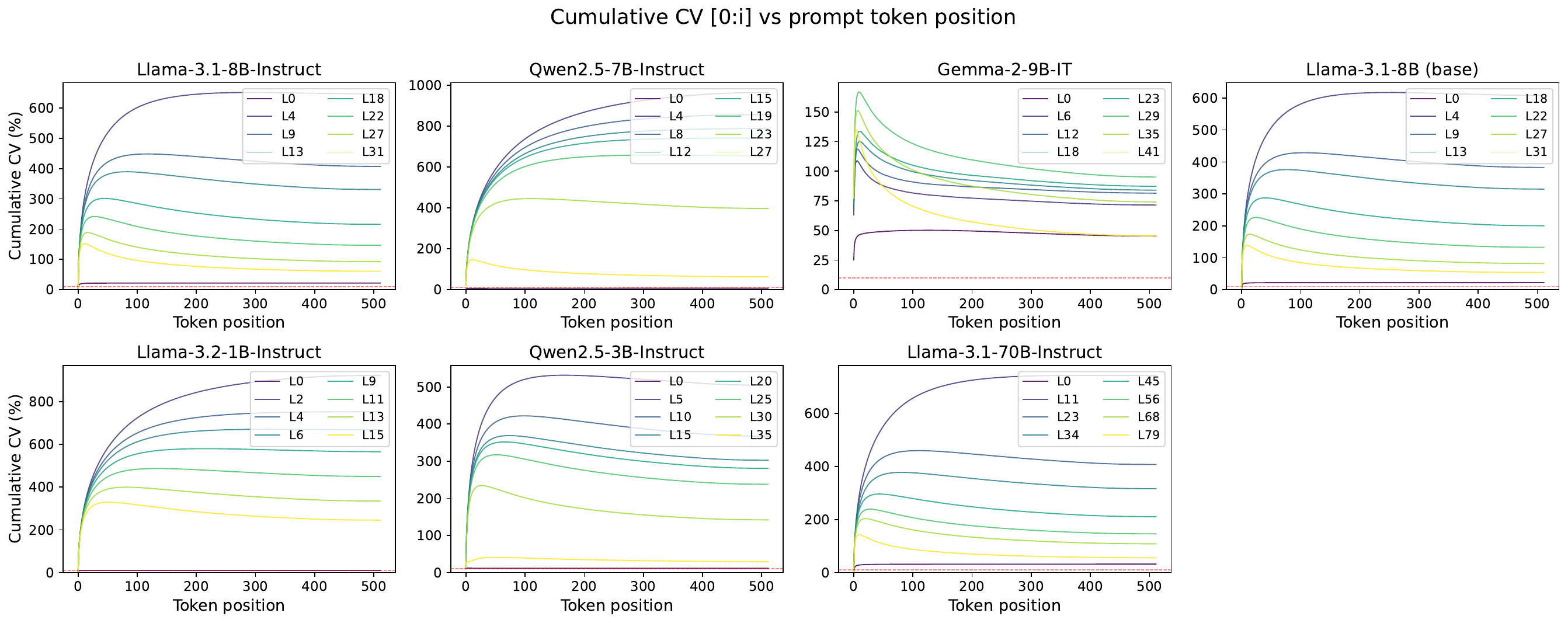}
\caption{
Cumulative CV over prompt-token positions. Pooling early attention-sink positions with later
content positions produces large prompt-token CV, showing that aggregate prompt statistics are
sensitive to position-dependent norm scale.
}
\label{fig:app_t1_cumulative_prompt}
\end{figure*}

\begin{figure*}[t]
\centering
\includegraphics[width=\textwidth]{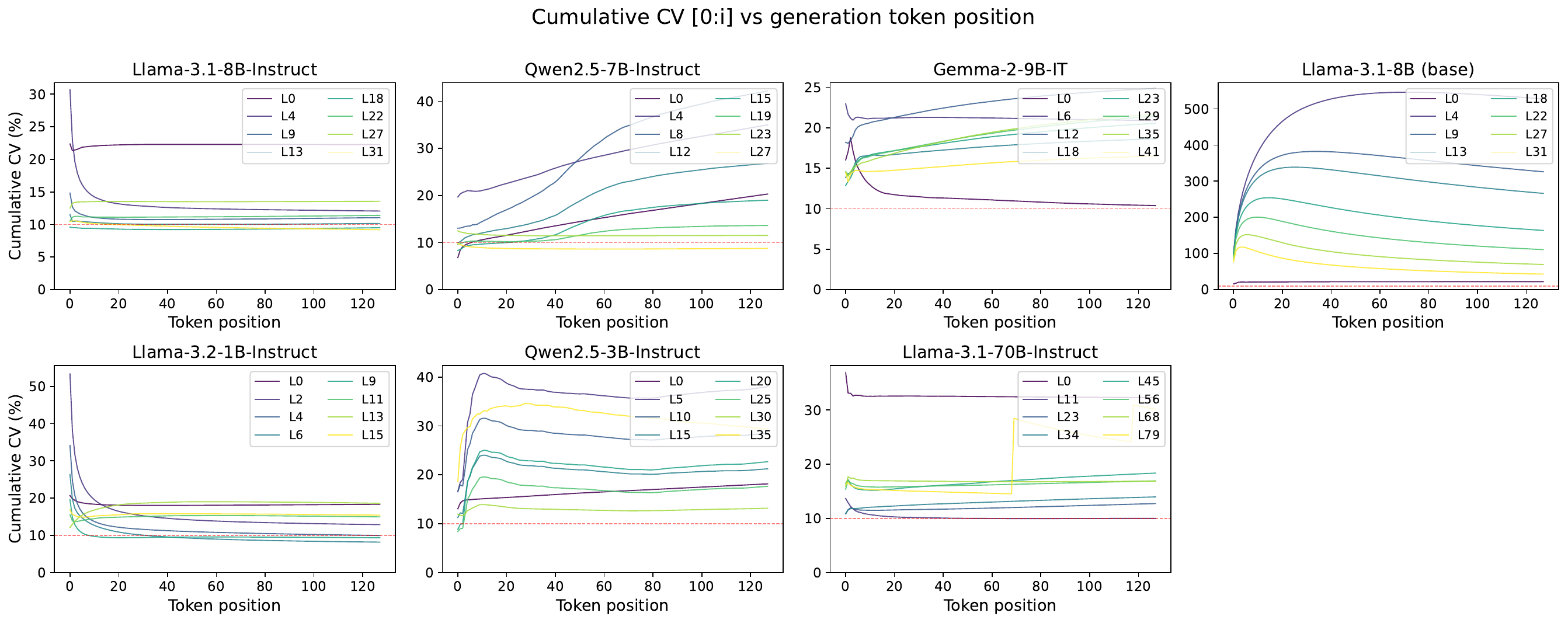}
\caption{
Cumulative CV over generation-token positions. The curves converge quickly for most
instruction-tuned models, indicating that generation-token norm variation is not dominated by a
small number of outlier positions.
}
\label{fig:app_t1_cumulative_gen}
\end{figure*}

\paragraph{Cumulative CV.}
Figures~\ref{fig:app_t1_cumulative_prompt} and~\ref{fig:app_t1_cumulative_gen} show cumulative
CV when positions are pooled from the start of the sequence. For prompt tokens, the early
attention-sink positions strongly affect the pooled statistic; as more content positions are included,
this effect is diluted. For generated tokens, the cumulative curves converge quickly, indicating that
generation-time norm variation is not dominated by a few outlier positions.

\begin{figure*}[t]
\centering
\includegraphics[width=\textwidth]{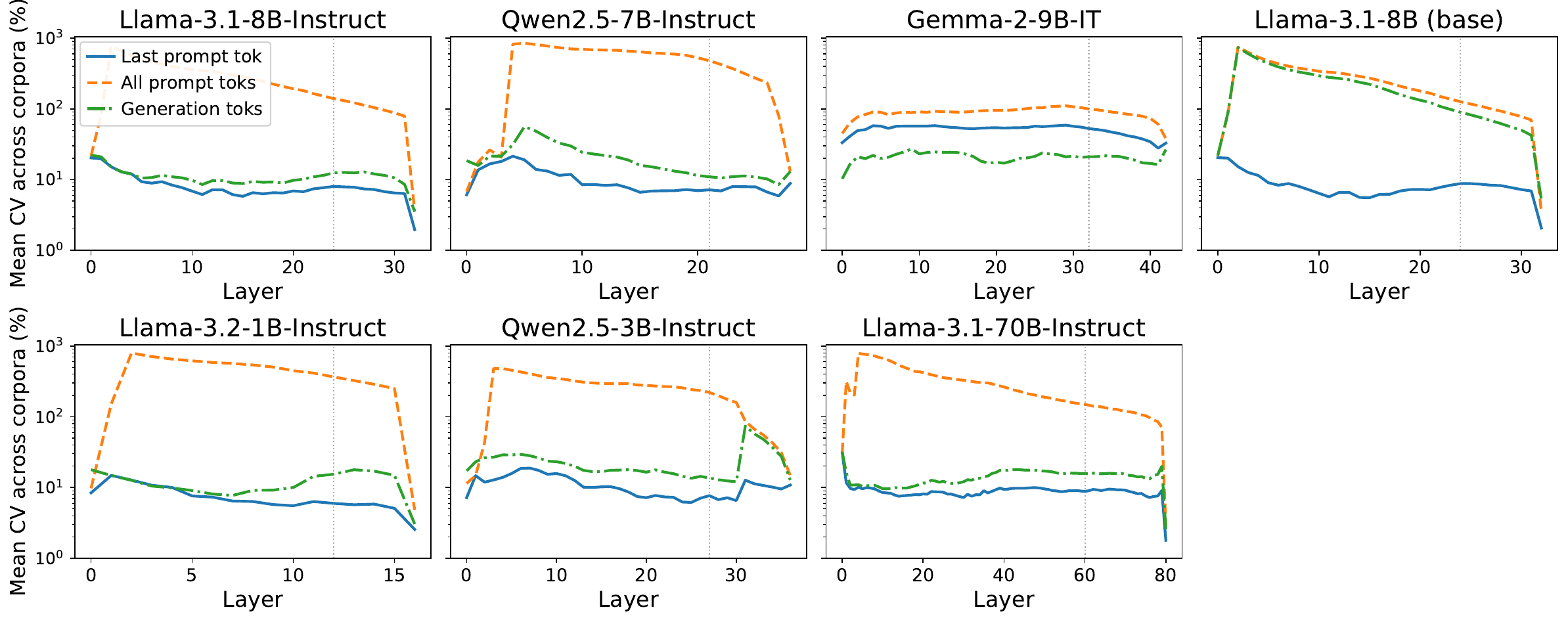}
\caption{
Mean CV across corpora for last prompt tokens, all prompt tokens, and generation tokens.
Position-dependent norm variation, especially from early attention-sink positions, strongly inflates
the all-prompt-token CV. Generation-token norms are more stable for most instruction-tuned
models.
}
\label{fig:app_t1_last_vs_all}
\end{figure*}

\paragraph{Layerwise token-population comparison.}
Figure~\ref{fig:app_t1_last_vs_all} compares the mean CV across corpora for three token
populations: the last prompt token, all prompt tokens, and generated tokens. The all-prompt-token
curve is much larger because it pools positions with different typical norm scales. In contrast,
generation-token CV is more stable for most instruction-tuned models, which supports the
interpretation that decoding-time steering operates on a comparatively stable radial landscape.

\begin{figure*}[t]
\centering
\includegraphics[width=\textwidth]{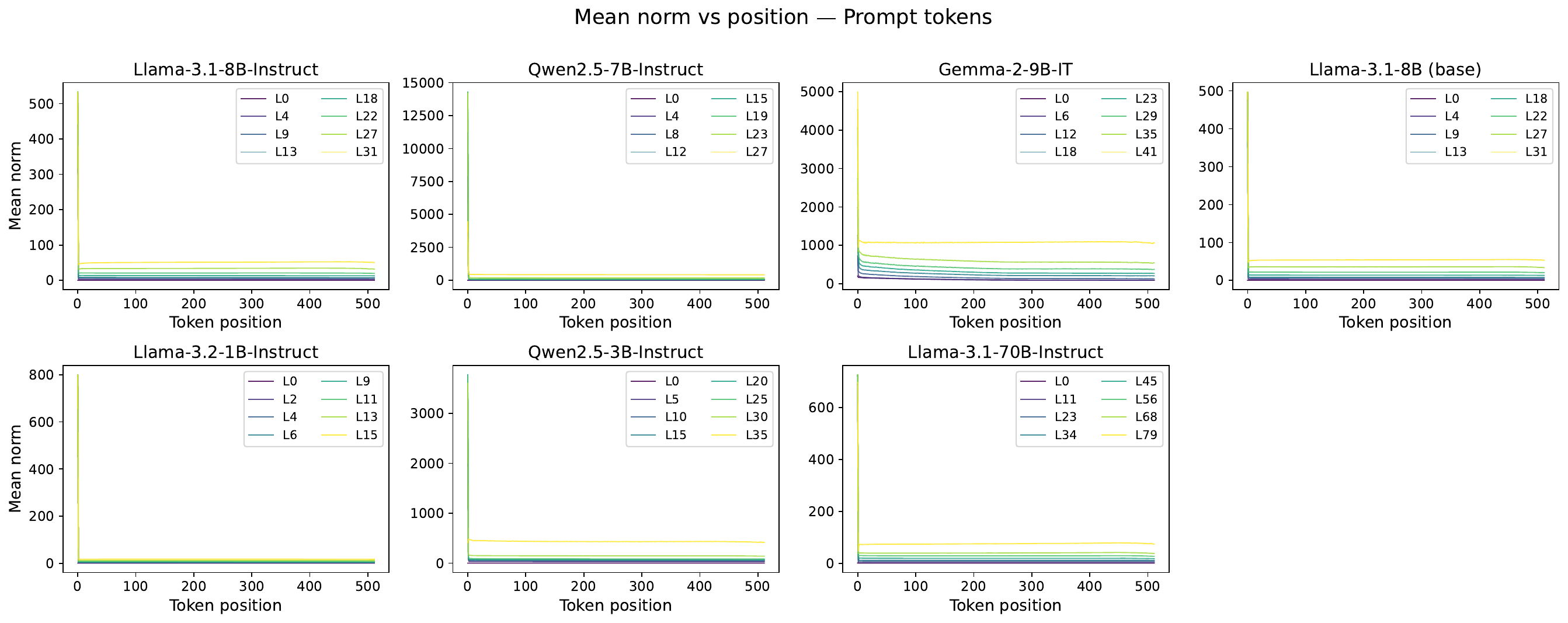}
\caption{
Mean hidden-state norm across prompt-token positions. Norms increase with layer depth, and the
first prompt position can have a disproportionately large norm in Llama and Qwen architectures.
}
\label{fig:app_t1_norm_prompt}
\end{figure*}

\begin{figure*}[t]
\centering
\includegraphics[width=\textwidth]{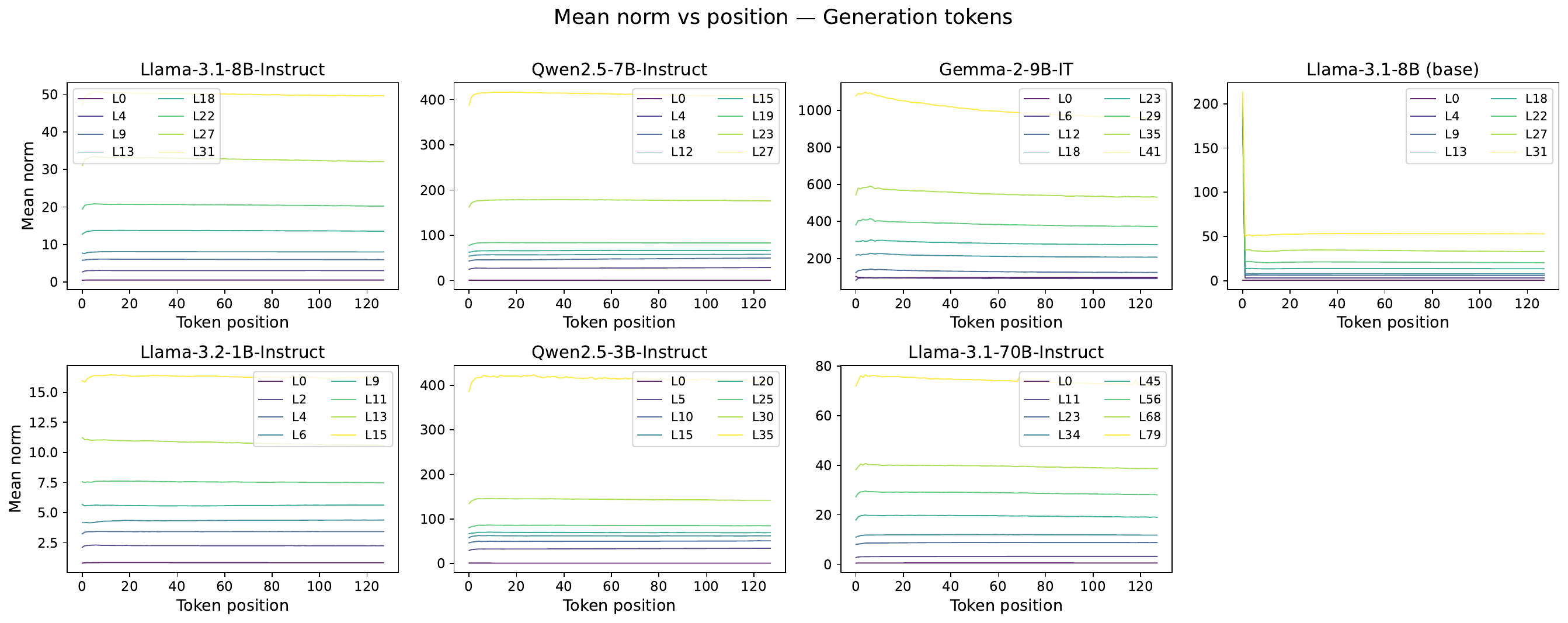}
\caption{
Mean hidden-state norm across generation-token positions. At each layer, generation-token norms
are nearly constant across positions for most instruction-tuned models.
}
\label{fig:app_t1_norm_gen}
\end{figure*}

\paragraph{Mean norm profiles.}
Figures~\ref{fig:app_t1_norm_prompt} and~\ref{fig:app_t1_norm_gen} report the corresponding
mean norm profiles. Prompt-token norms show strong position effects, especially at the first token
in Llama and Qwen architectures. In contrast, generation-token norms are nearly constant across
positions at a fixed layer for most instruction-tuned models. This supports the main-text
interpretation that prompt-token statistics can be strongly position-dependent, while decoding
positions are more stable.

Overall, prompt-token norms are strongly affected by position, whereas generation-token norms are
more stable across decoding steps. Thus, norm preservation in spherical steering should be
understood as preserving each token's own radius, not as forcing all activations onto a shared
global radius.

\paragraph{Token-population summary.}
Table~\ref{tab:app_t1_token_populations} summarizes norm CV at the 75\%-depth layer for the
three token populations used in the norm-variation analysis. The all-prompt-token statistic is much
larger because it pools positions with very different typical norm scales. Generated tokens are more
stable for most instruction-tuned models, which is the regime most relevant to steering during
decoding.

\begin{table*}[t]
\centering
\small
\caption{
Norm CV at the 75\%-depth layer for three token populations, averaged across corpora.
Generation tokens are the positions directly modified by the steering hook during decoding.
}
\label{tab:app_t1_token_populations}
\resizebox{\textwidth}{!}{
\begin{tabular}{lccc}
\toprule
Model & Last prompt token & All prompt tokens & Generation tokens \\
\midrule
Llama-3.1-8B-Instruct  & 8.0\%  & 140.0\% & 12.5\% \\
Qwen2.5-7B-Instruct    & 7.2\%  & 478.8\% & 11.0\% \\
Gemma-2-9B-it          & 52.7\% & 99.6\%  & 21.1\% \\
Llama-3.1-8B           & 8.8\%  & 126.1\% & 89.6\% \\
Llama-3.2-1B-Instruct  & 6.0\%  & 365.3\% & 15.4\% \\
Qwen2.5-3B-Instruct    & 7.7\%  & 220.6\% & 13.7\% \\
Llama-3.1-70B-Instruct & 8.8\%  & 149.6\% & 15.7\% \\
\bottomrule
\end{tabular}
}
\end{table*}

\section{Additional Directional-Encoding Results}
\label{app:directional_encoding}

The main text reports the layerwise probe curves showing that concept information is primarily
encoded in activation direction. Here we provide the corresponding per-model and per-dataset probe
accuracies. We compare linear probes trained on raw hidden states, unit-normalized hidden states,
and norm-only features. Across all evaluated concepts, unit-normalized probes closely match raw
probes, while norm-only probes remain near chance.

\begin{table*}[t]
\centering
\scriptsize
\caption{
Linear-probe accuracies for raw, unit-normalized, and norm-only representations. Unit-normalized
features retain almost all of the predictive information in raw hidden states, while norm-only
features remain close to chance.
}
\label{tab:app_t2_probe_acc}
\resizebox{\textwidth}{!}{
\begin{tabular}{llccc}
\toprule
Model & Dataset & Raw $h$ & Unit $h/\|h\|$ & Norm-only $\|h\|$ \\
\midrule
Llama-3.1-8B-Instruct & TruthfulQA      & 81.6\% & 81.1\% & 53.8\% \\
Llama-3.1-8B-Instruct & SST-2           & 92.5\% & 92.4\% & 53.4\% \\
Llama-3.1-8B-Instruct & CivilComments   & 84.0\% & 83.9\% & 53.1\% \\
Llama-3.1-8B-Instruct & IMDB            & 88.5\% & 88.4\% & 47.9\% \\
\midrule
Qwen2.5-7B-Instruct & TruthfulQA        & 70.1\% & 70.1\% & 57.9\% \\
Qwen2.5-7B-Instruct & SST-2             & 91.5\% & 91.4\% & 51.0\% \\
Qwen2.5-7B-Instruct & CivilComments     & 80.6\% & 80.2\% & 52.9\% \\
Qwen2.5-7B-Instruct & IMDB              & 87.3\% & 86.9\% & 53.7\% \\
\midrule
Gemma-2-9B-it & TruthfulQA             & 57.6\% & 58.4\% & 51.7\% \\
Gemma-2-9B-it & SST-2                  & 71.6\% & 71.0\% & 48.8\% \\
Gemma-2-9B-it & CivilComments          & 65.9\% & 65.0\% & 46.6\% \\
Gemma-2-9B-it & IMDB                   & 75.4\% & 74.5\% & 51.1\% \\
\midrule
Llama-3.1-8B & TruthfulQA              & 80.4\% & 80.6\% & 51.5\% \\
Llama-3.1-8B & SST-2                   & 92.8\% & 93.0\% & 49.7\% \\
Llama-3.1-8B & CivilComments           & 84.0\% & 84.0\% & 56.8\% \\
Llama-3.1-8B & IMDB                    & 88.9\% & 88.9\% & 53.9\% \\
\midrule
Llama-3.2-1B-Instruct & TruthfulQA       & 73.6\% & 74.0\% & 56.2\% \\
Llama-3.2-1B-Instruct & SST-2            & 88.9\% & 88.9\% & 51.0\% \\
Llama-3.2-1B-Instruct & CivilComments    & 78.6\% & 78.4\% & 48.1\% \\
Llama-3.2-1B-Instruct & IMDB             & 83.6\% & 83.8\% & 51.0\% \\
\midrule
Qwen2.5-3B-Instruct & TruthfulQA         & 66.2\% & 66.5\% & 56.1\% \\
Qwen2.5-3B-Instruct & SST-2              & 87.5\% & 87.5\% & 52.9\% \\
Qwen2.5-3B-Instruct & CivilComments      & 76.1\% & 76.5\% & 54.6\% \\
Qwen2.5-3B-Instruct & IMDB               & 79.5\% & 79.9\% & 55.1\% \\
\midrule
Llama-3.1-70B-Instruct & TruthfulQA       & 86.0\% & 85.5\% & 50.5\% \\
Llama-3.1-70B-Instruct & SST-2            & 92.0\% & 92.5\% & 45.2\% \\
Llama-3.1-70B-Instruct & CivilComments    & 83.9\% & 84.5\% & 47.1\% \\
Llama-3.1-70B-Instruct & IMDB             & 91.2\% & 91.0\% & 52.4\% \\
\bottomrule
\end{tabular}
}
\end{table*}

The table supports the directional-encoding claim used throughout the paper. Normalizing hidden
states to unit length causes almost no loss in probe accuracy, indicating that the concepts remain
linearly accessible after removing the radial component. In contrast, probes trained only on the
activation norm are close to chance for all datasets and model families. This pattern also holds for
Gemma, where norm variation is much larger than in the Llama and Qwen models, showing that
large radial variability does not imply that the concept itself is encoded in the norm.

\section{CAA-m Per-Token Matching Algorithm}
\label{app:caam}

CAA-m chooses a separate additive coefficient for every token so that the normalized output reaches
the requested concept score. Let
\[
    x = r(c s+\sqrt{1-c^2}\,v),
\]
where $r=\|x\|$, $c=\langle x/\|x\|,s\rangle$, and $v$ is the unit component of $x/\|x\|$ orthogonal to $s$.
For $y=x+\alpha s$, the target constraint is
\[
    \left\langle \frac{y}{\|y\|},s \right\rangle = \gamma.
\]
Since
\[
    y = (rc+\alpha)s + r\sqrt{1-c^2}\,v,
\]
solving the constraint gives
\begin{equation}
    \alpha
    =
    r\left(
    \frac{\gamma\sqrt{1-c^2}}{\sqrt{1-\gamma^2}}
    -
    c
    \right).
    \label{eq:app-caam-closed-form}
\end{equation}
This expression is well-defined for $\gamma \in (-1,1)$. When $|\gamma|$ approaches 1, the required additive coefficient can become large, reflecting the fact that an almost perfectly aligned target direction may require a large displacement for tokens whose initial residual component orthogonal to $s$ is large. Thus it controls the angular concept score while allowing the norm to change.

\section{Additional Fixed-Angle Steering Results}
\label{app:t3a_s_vs_caam}

This appendix provides additional results for the comparison between S and CAA-m at matched
per-token target $\gamma$. Since both methods reach the same normalized concept direction, their
difference is radial: S preserves the original norm, while CAA-m leaves the additive norm change
intact. Here we show per-dataset gaps comparison and the full
per-cell tables.

\begin{figure*}[t]
\centering
\includegraphics[width=\textwidth]{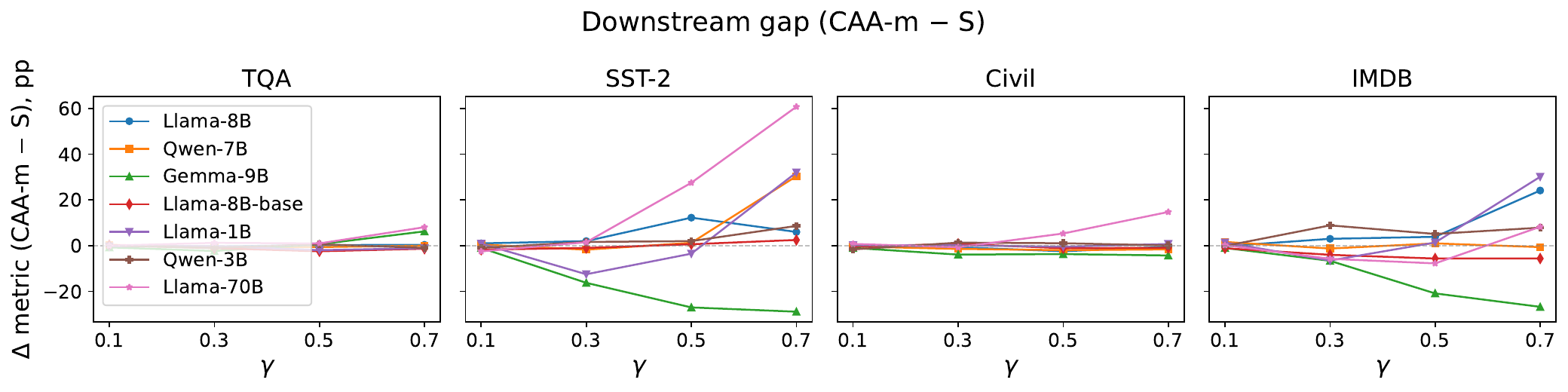}\\[6pt]
\includegraphics[width=\textwidth]{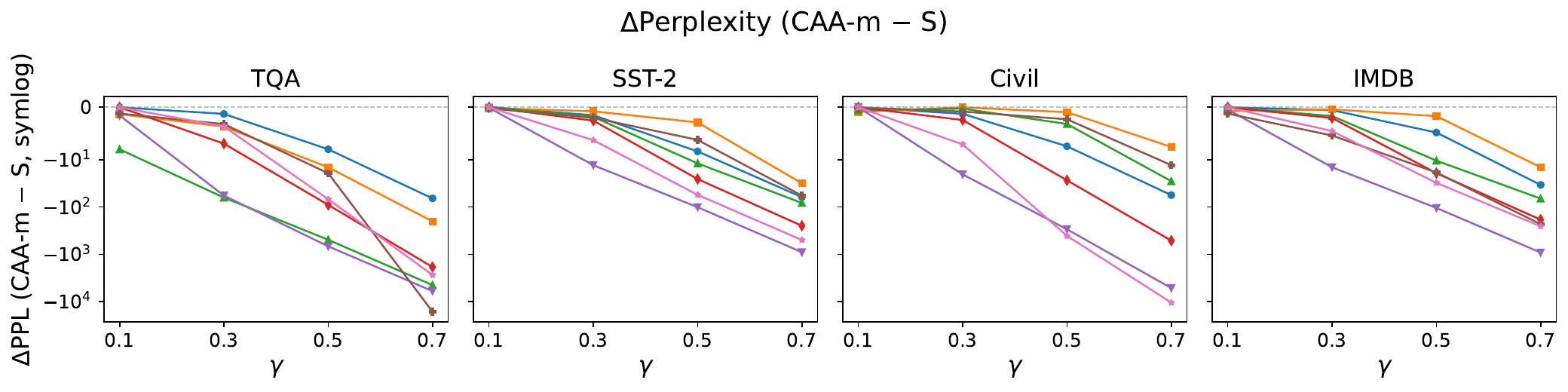}\\[6pt]
\includegraphics[width=\textwidth]{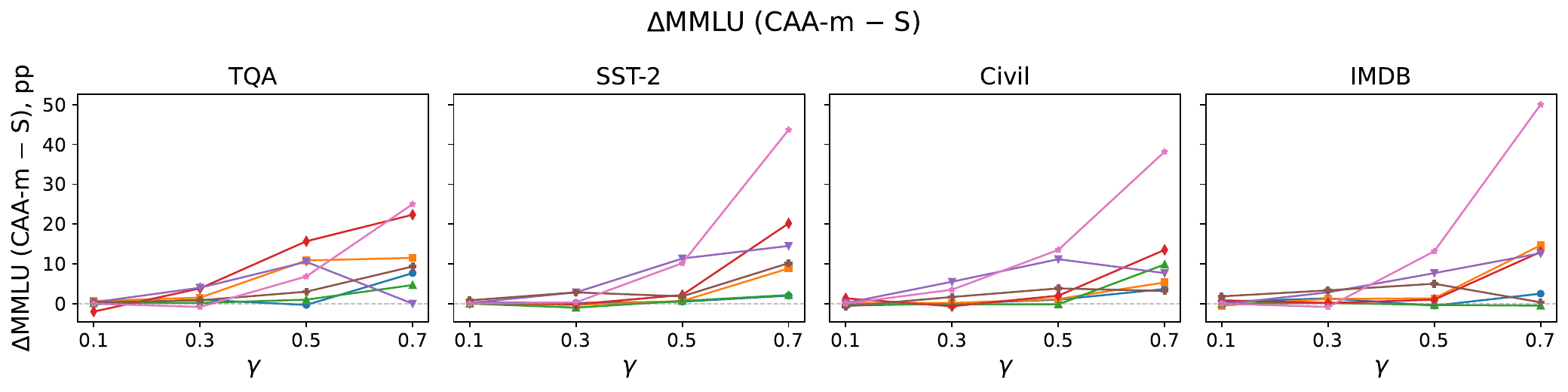}
\caption{
Per-dataset S vs.\ CAA-m gaps at matched per-token target $\gamma$.
Top: downstream-metric gap, CAA-m $-$ S, in percentage points.
Middle: WikiText-103 perplexity gap, CAA-m $-$ S, using a symlog scale.
Bottom: MMLU gap, CAA-m $-$ S, in percentage points.
The dashed grey line marks zero gap. For downstream metrics and MMLU, positive values favour
CAA-m. For perplexity, negative values favour CAA-m because lower perplexity is better.
}
\label{fig:app_t3a_svscaa_delta}
\end{figure*}

\paragraph{Per-dataset gaps.}
Figure~\ref{fig:app_t3a_svscaa_delta} reports the difference between CAA-m and S across
datasets and models. At low $\gamma$, the two methods are close on all metrics. At larger
$\gamma$, CAA-m opens a large stability gap: it usually has much lower perplexity and higher
MMLU accuracy, while the downstream task metric remains comparable on average but varies more
across models and datasets.

\section{Additional Fixed-Strength Steering Results}
\label{app:t3b_fixed_strength}

This appendix provides additional results for the fixed-strength methods: CAA, CAA-r, and AS.
Unlike S and CAA-m, these methods use a single global steering parameter and are calibrated to
match the target mean concept score $\bar\gamma$. This comparison isolates whether norm
preservation alone explains downstream stability. CAA-r and AS both preserve the hidden-state
norm, while CAA does not; however, the results show that the token-level angular profile is more
important than norm preservation alone.

\begin{figure*}[t]
\centering
\includegraphics[width=\textwidth]{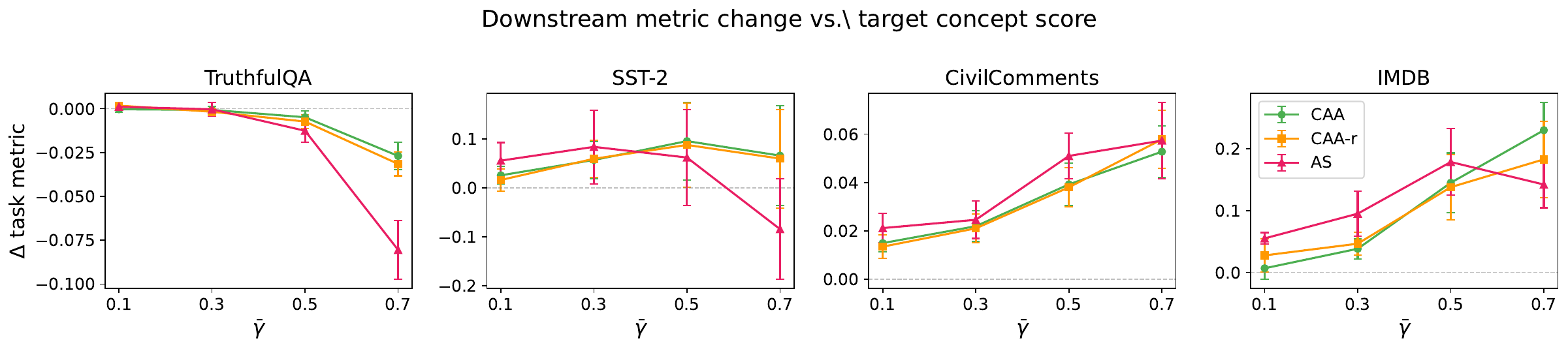}
\caption{
Mean downstream metric change, $\Delta$ task, versus target mean concept score, averaged across
models per dataset. CAA, CAA-r, and AS produce similar gains at moderate targets, while AS
diverges at high $\bar\gamma$ because its fixed spherical displacement causes larger token-level
disruption.
}
\label{fig:app_t3b_downstream_vs_gamma}
\end{figure*}

\paragraph{Downstream trajectory.}
Figure~\ref{fig:app_t3b_downstream_vs_gamma} compares the downstream-metric trajectory of the
three fixed-strength methods as the target mean concept score increases. The methods behave
similarly at moderate targets, while AS becomes less stable at high $\bar\gamma$.

\begin{figure*}[t]
\centering
\includegraphics[width=\textwidth]{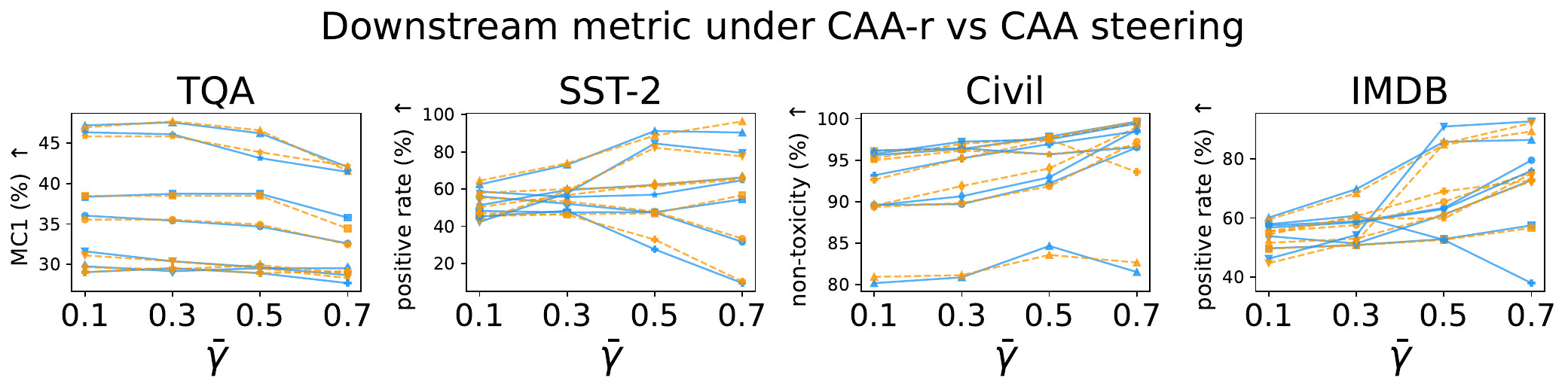}\\[6pt]
\includegraphics[width=\textwidth]{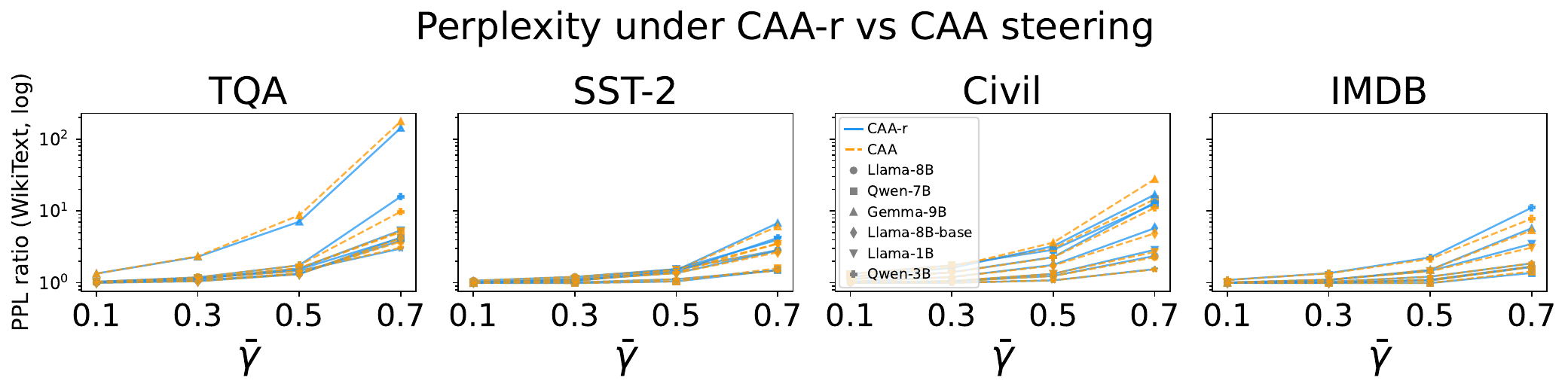}
\caption{
CAA-r versus CAA at matched mean concept score. Top: downstream metric versus
$\bar{\gamma}$. Bottom: WikiText-103 PPL ratio. Since CAA-r only renormalizes the additive CAA
output, the two methods remain close in downstream behavior.
}
\label{fig:app_t3b_caar_vs_caa}
\end{figure*}

\begin{figure*}[t]
\centering
\includegraphics[width=\textwidth]{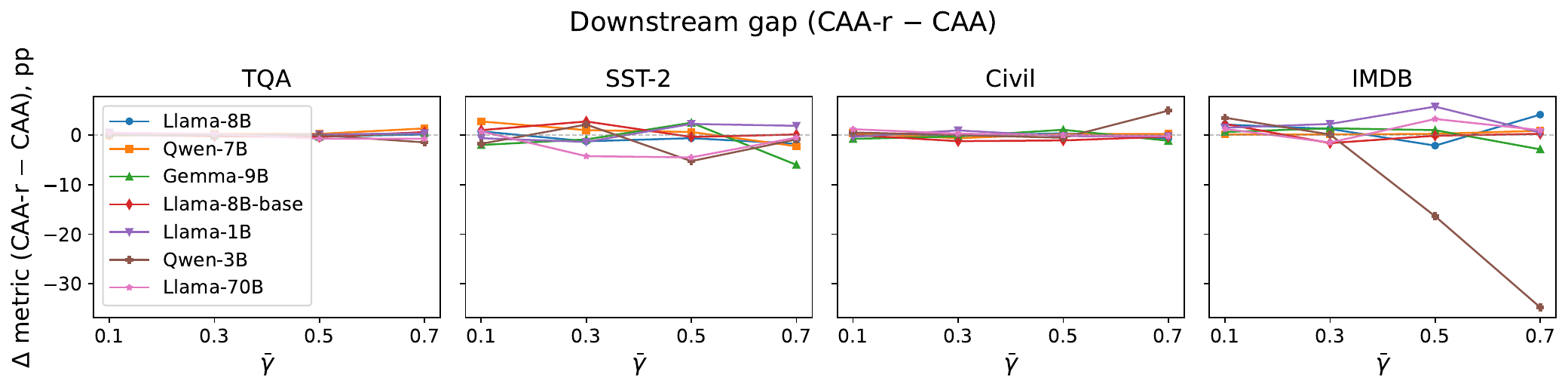}\\[6pt]
\includegraphics[width=\textwidth]{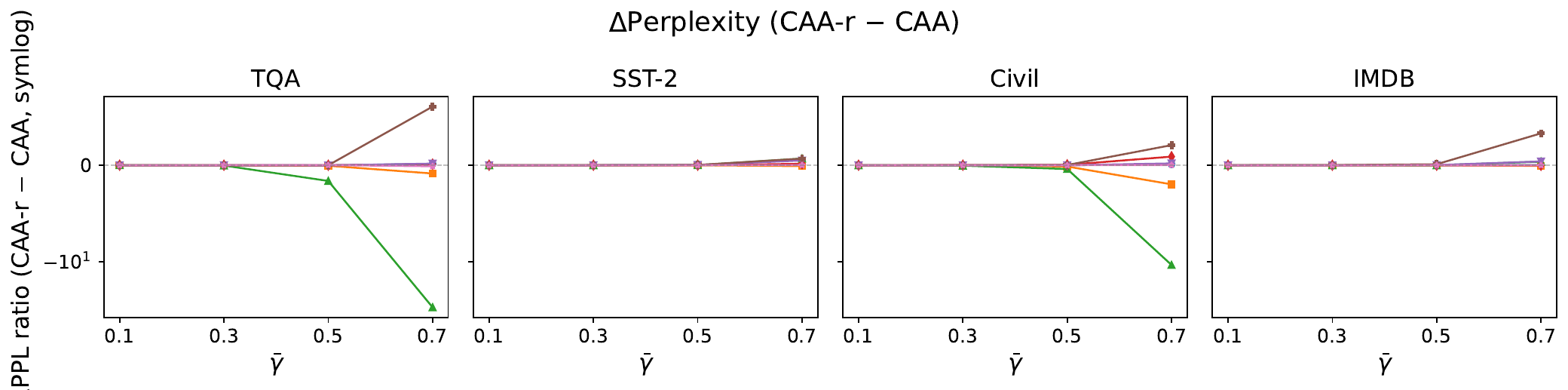}
\caption{
CAA-r $-$ CAA gap per dataset, with one line per model. Top: downstream-metric difference in
percentage points. Bottom: WikiText-103 PPL-ratio difference, shown on a symlog scale. The dashed
grey line marks zero gap. The gaps remain small across most targets, showing that renormalizing
CAA does not substantially change behavior in this fixed-strength regime.
}
\label{fig:app_t3b_caar_vs_caa_delta}
\end{figure*}

\paragraph{CAA-r versus CAA.}
Figure~\ref{fig:app_t3b_caar_vs_caa} compares CAA-r and CAA at matched mean concept score.
These methods have the same normalized output direction after the additive update; CAA-r only
rescales the result back to the original norm. Consequently, their downstream and PPL curves stay
close across targets. Figure~\ref{fig:app_t3b_caar_vs_caa_delta} shows the same comparison as
per-dataset gaps, confirming that post-hoc renormalization is not the main factor controlling
stability in the fixed-strength setting.

\begin{figure*}[t]
\centering
\includegraphics[width=\textwidth]{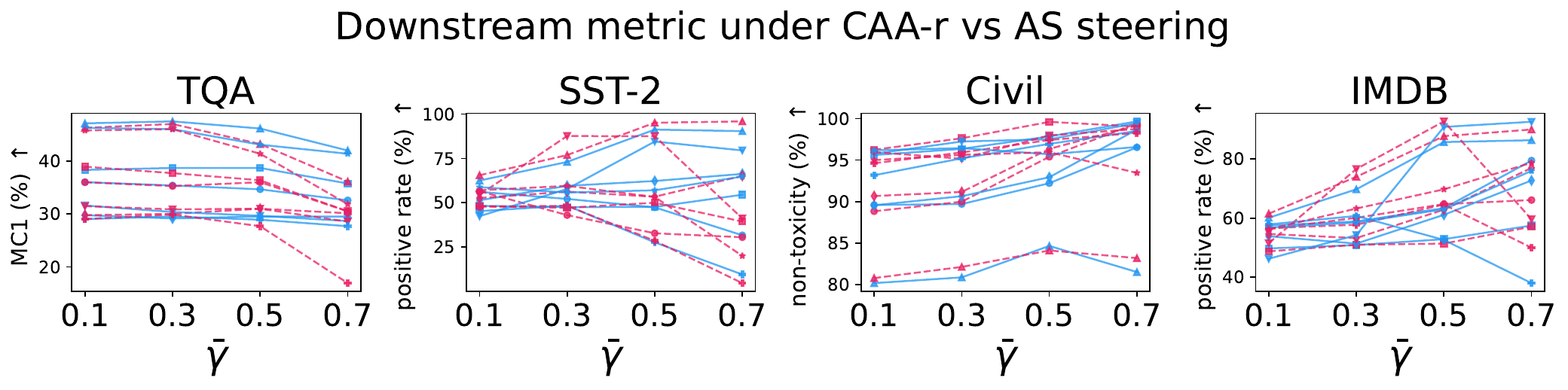}\\[6pt]
\includegraphics[width=\textwidth]{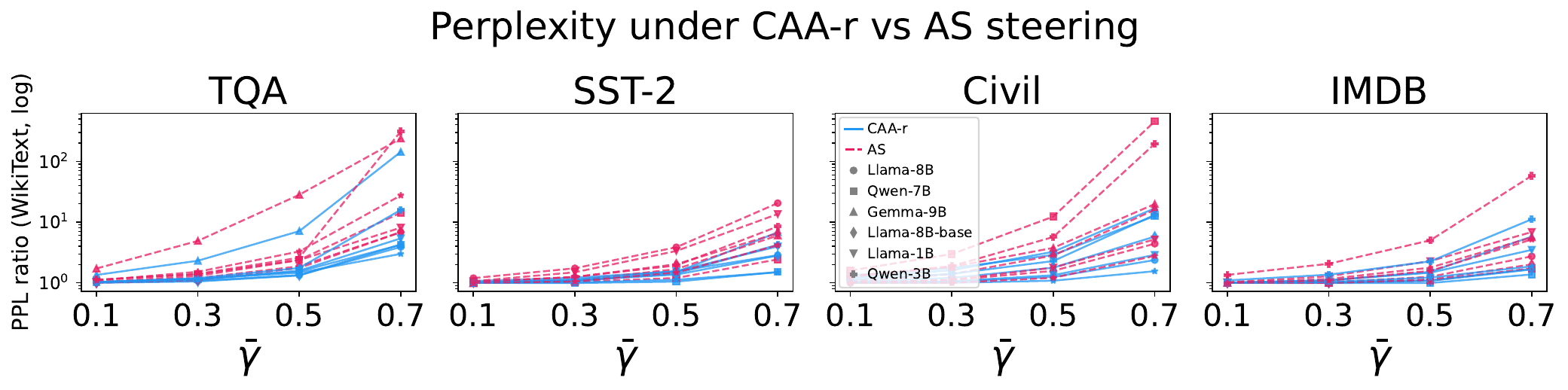}
\caption{
CAA-r versus AS at matched mean concept score. Top: downstream metric versus
$\bar{\gamma}$. Bottom: WikiText-103 PPL ratio. Both methods preserve the hidden-state norm,
but they induce different token-level angular profiles. AS becomes much more costly in PPL at high
$\bar{\gamma}$, showing that norm preservation alone is not sufficient for stable steering.
}
\label{fig:app_t3b_caar_vs_as}
\end{figure*}

\begin{figure*}[t]
\centering
\includegraphics[width=\textwidth]{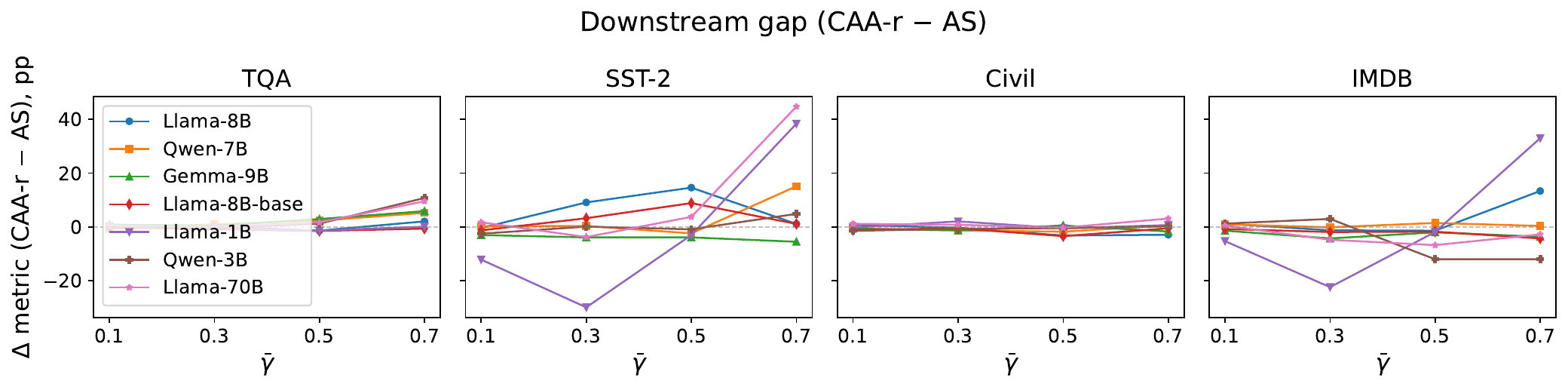}\\[6pt]
\includegraphics[width=\textwidth]{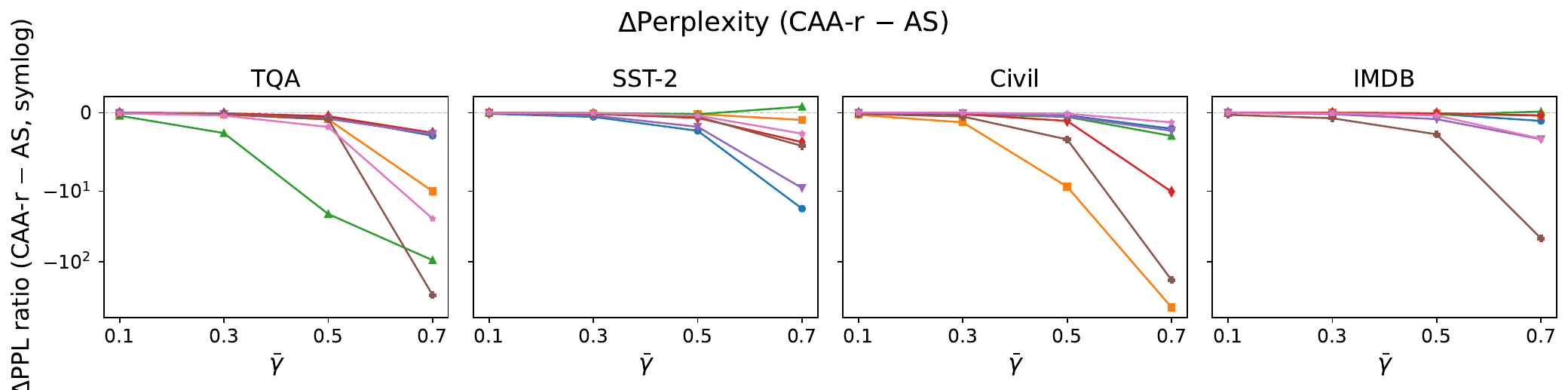}
\caption{
CAA-r $-$ AS gap per dataset, with one line per model. Top: downstream-metric difference in
percentage points. Bottom: WikiText-103 PPL-ratio difference, shown on a symlog scale. Negative
PPL gaps mean CAA-r has lower perplexity than AS. Although both methods preserve norm, AS
incurs much larger PPL degradation at high $\bar{\gamma}$.
}
\label{fig:app_t3b_caar_vs_as_delta}
\end{figure*}

\paragraph{CAA-r versus AS.}
Figure~\ref{fig:app_t3b_caar_vs_as} compares CAA-r and AS at matched mean concept score.
Both methods preserve $\|y\|=\|x\|$, but they distribute the angular intervention differently across
tokens. CAA-r inherits a token-dependent angular displacement from the additive update, whereas
AS applies a fixed spherical displacement. Figure~\ref{fig:app_t3b_caar_vs_as_delta} shows that
this difference produces a large PPL gap at high $\bar{\gamma}$: AS becomes substantially less
stable despite preserving the norm. Thus, norm preservation alone is not sufficient to explain the
steering--quality trade-off.

\begin{figure*}[t]
\centering
\includegraphics[width=0.48\textwidth]{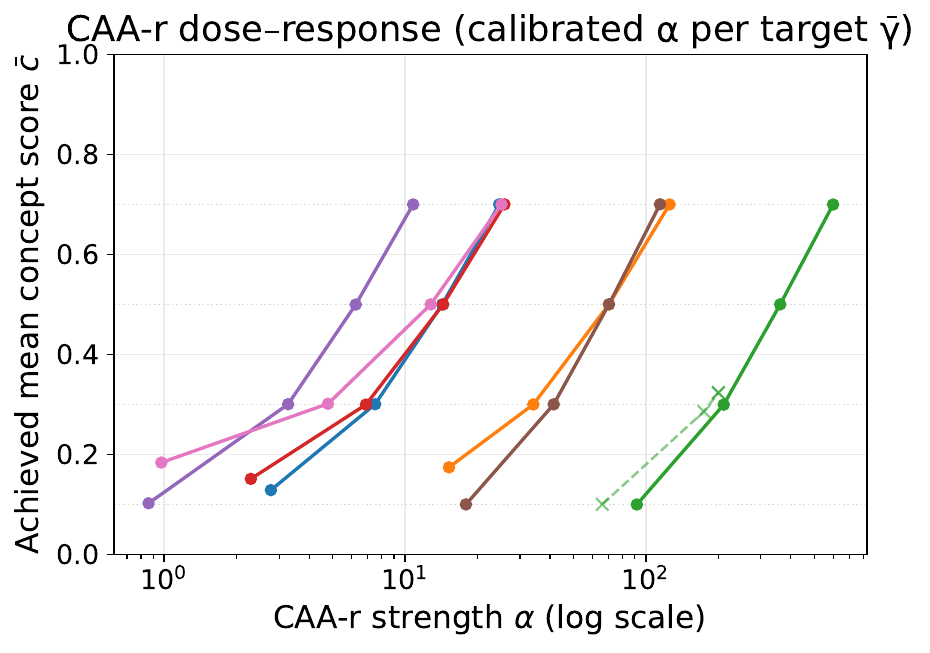}
\hfill
\includegraphics[width=0.48\textwidth]{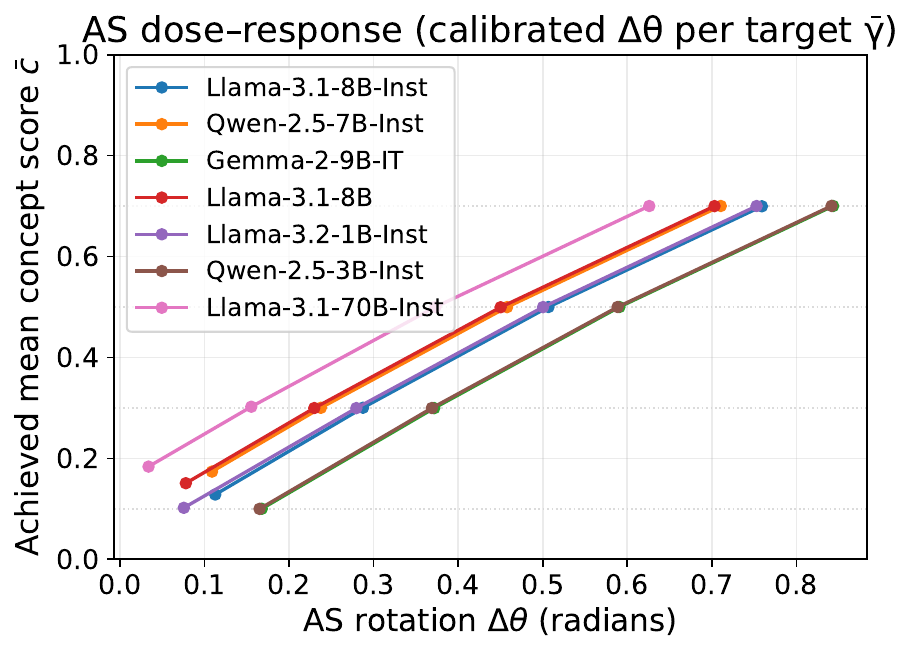}
\caption{
Dose-response curves for fixed-strength calibration. Left: CAA-r mean concept score versus
additive strength $\alpha$ on a log scale. Right: AS mean concept score versus angular displacement
$\Delta\theta$. CAA-r calibration is sensitive to residual-stream norm scale, while AS calibration is
norm-invariant.
}
\label{fig:app_t3b_dose}
\end{figure*}

\paragraph{Calibration dose response.}
Figure~\ref{fig:app_t3b_dose} shows the calibration curves used to match the target mean concept
score. For CAA-r, the required additive strength is highly model-dependent because it depends on
the scale of the residual stream. Gemma requires a much wider search range for $\alpha$. In
contrast, AS is calibrated by angular displacement and is therefore less sensitive to activation norm
scale.

\section{Concept-Score Closure}
\label{app:concept_score_closure}

This appendix compares the steering methods by how tightly they close the gap to the requested concept score. The main experiments compare downstream behavior and perplexity; here we isolate the intervention itself by measuring the achieved per-token concept score after steering. This diagnostic separates methods that enforce a target score token-by-token from methods that only
match a target score on average.

\begin{figure*}[t]
\centering
\includegraphics[width=\textwidth]{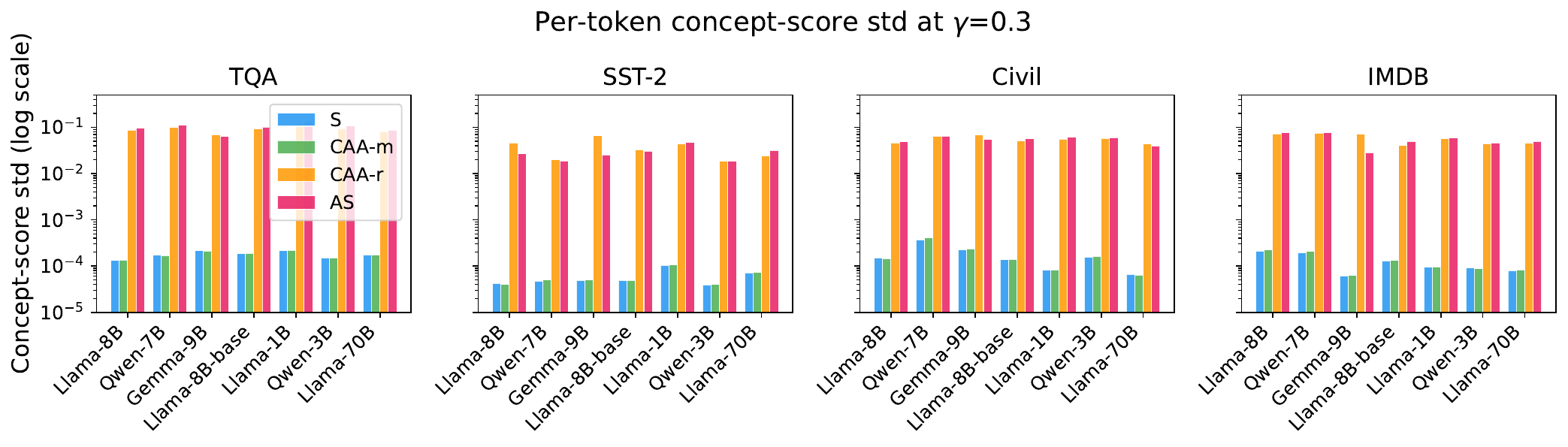}
\caption{
Per-token concept-score standard deviation at matched target score. Per-token targeted methods
collapse tightly around the requested value, while fixed-strength methods have much larger spread.
This shows that matching the mean concept score is not equivalent to closing the concept score for
each token.
}
\label{fig:app_t6_std}
\end{figure*}

\paragraph{Per-token score variance.}
Figure~\ref{fig:app_t6_std} reports the standard deviation of achieved concept scores across
tokens at the target level used in the closure diagnostic. S and CAA-m have near-zero spread because
they explicitly solve for the target concept score for each token. In contrast, CAA, CAA-r, and AS
use a single global steering strength. Even when their mean achieved score is calibrated to the
target, individual tokens spread over a much wider range. This confirms that concept-score closure
is a separate axis of method design: two methods can have the same average steering strength but
very different token-level precision.

\begin{figure*}[t]
\centering
\includegraphics[width=\textwidth]{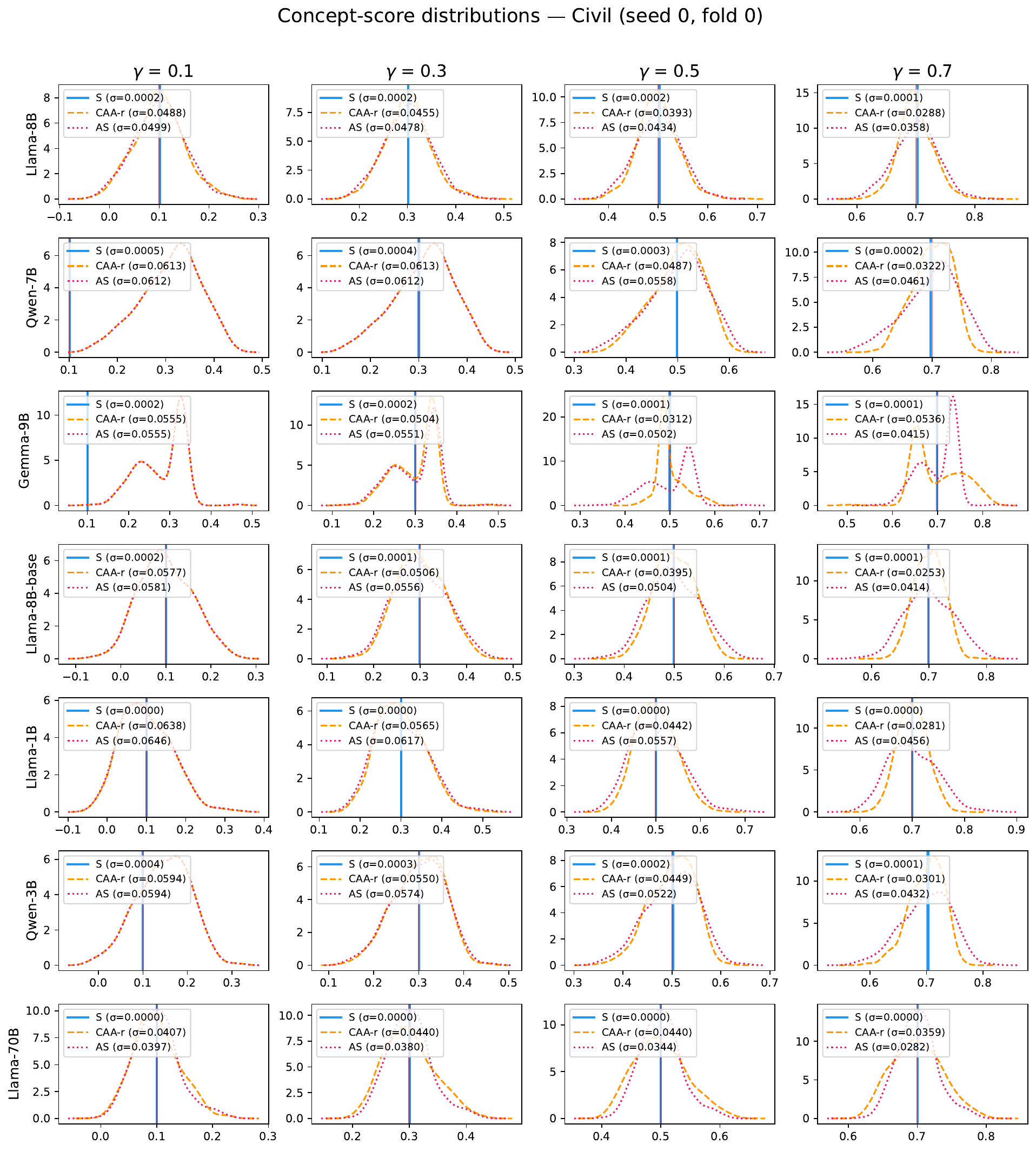}
\caption{
Achieved concept-score distributions on CivilComments. Each panel corresponds to one model and
target score. Targeted methods collapse near the requested score, while fixed-strength methods
spread across a wider interval even when calibrated to the same mean. At higher target scores, the
CAA-r and AS distributions become more different, reflecting their distinct token-level angular
profiles.
}
\label{fig:app_t6_kde_civil}
\end{figure*}

\paragraph{Concept-score distributions.}
Figure~\ref{fig:app_t6_kde_civil} shows the full achieved-score distributions on CivilComments.
The distributional view makes the same point as the standard-deviation summary: targeted methods
produce a sharp peak at the requested score, whereas fixed-strength methods produce broader
token-level distributions. The difference between CAA-r and AS also becomes more pronounced at
higher target scores. Although both methods are calibrated to the same mean concept score and both
preserve the hidden-state norm, their achieved-score distributions diverge at large $\gamma$ because
they induce different token-level angular profiles.

Overall, these results justify separating \emph{semantic strength} from \emph{concept-score
closure}. Mean-matched fixed-strength methods can express the target concept on average, but they
do not apply the same intervention to every token. Per-token targeted methods close the concept
score much more precisely, which explains why they occupy a distinct part of the control--quality
trade-off in the main experiments.

\section{Off-Arc Perturbations}
\label{app:off_arc}

This appendix tests whether the great-circle arc used by S is empirically meaningful, not only
geometrically minimal. Starting from the spherical solution, we perturb the residual component
away from the arc while keeping both the hidden-state norm and the target concept score fixed.
Thus, any degradation caused by the perturbation cannot be explained by weaker concept control
or by a different norm; it must come from moving away from the task-relevant residual direction.

Using the notation from the main text, we perturb the residual direction by an angle $\delta$ toward
a direction $q$ orthogonal to both the concept direction and the original residual direction:
\[
    y(\delta)
    =
    \|x\|
    \left(
        \gamma s
        +
        \sqrt{1-\gamma^2}
        \left(
            \cos\delta \, v
            +
            \sin\delta \, q
        \right)
    \right).
\]
All points on this sweep have the same norm and the same concept score. The arc solution is
$\delta=0$. If the great-circle arc is the empirically relevant axis, then perturbing in either
direction should degrade perplexity, MMLU, or downstream task performance.

\paragraph{Aggregate off-arc degradation.}
Table~\ref{tab:app_t5_aggregate} reports degradation relative to the arc solution. PPL increase away from $\delta=0$, while MMLU and downstream task metrics generally decrease.
The effect is approximately symmetric and becomes stronger as $|\delta|$ increases.

\begin{table}[t]
\centering
\small
\caption{
Aggregate degradation under off-arc perturbations. PPL is reported as ratios relative to
$\delta=0$; MMLU and downstream metrics are reported as absolute changes relative to
$\delta=0$. PPL ratio above $1$ indicate degradation, while negative MMLU/downstream
changes indicate degradation.
}
\label{tab:app_t5_aggregate}
\begin{tabular}{lcccc}
\toprule
Metric & $\delta=-0.2$ & $\delta=-0.1$ & $\delta=+0.1$ & $\delta=+0.2$ \\
\midrule
PPL ratio $(\delta / 0)$ & $\times 1.22$ & $\times 1.05$ & $\times 1.06$ & $\times 1.23$ \\
$\Delta$MMLU            & $-0.014$ & $-0.003$ & $-0.003$ & $-0.012$ \\
$\Delta$Downstream      & $-0.005$ & $+0.001$ & $-0.004$ & $-0.013$ \\
\bottomrule
\end{tabular}
\end{table}

\paragraph{Perturbation direction type.}
Table~\ref{tab:app_t5_direction_type} breaks the PPL effect down by the type of off-arc direction.
Random directions produce the mildest degradation, PCA directions produce the steepest valleys,
and cross-dataset directions fall in between. This suggests that the most important residual
directions are aligned with high-variance structure in the residual subspace, while concept axes from
related datasets also overlap with task-relevant residual variation.

\begin{table}[t]
\centering
\small
\caption{
PPL ratio by off-arc perturbation direction type, averaged across completed cells. PCA directions
produce the steepest degradation, consistent with the residual subspace containing task-relevant
variation.
}
\label{tab:app_t5_direction_type}
\begin{tabular}{lcccc}
\toprule
Direction type & $\delta=-0.2$ & $\delta=-0.1$ & $\delta=+0.1$ & $\delta=+0.2$ \\
\midrule
Random $(n=2)$        & $\times 1.15$ & $\times 1.03$ & $\times 1.06$ & $\times 1.21$ \\
PCA $(n=2)$           & $\times 1.43$ & $\times 1.09$ & $\times 1.07$ & $\times 1.32$ \\
Cross-dataset $(n=4)$ & $\times 1.20$ & $\times 1.05$ & $\times 1.06$ & $\times 1.23$ \\
\bottomrule
\end{tabular}
\end{table}

PPL is minimized at $\delta=0$ in almost all completed cells, and the few exceptions have negligible relative gaps.
Overall, perturbing away from the spherical arc worsens model behavior even though the concept
score and norm are held fixed. This supports the interpretation that the S direction is not merely
the shortest geometric edit, but also the empirically task-relevant residual direction.

\section{Additional Norm-Scaling Results}
\label{app:t4_beta_sweep}

This appendix provides the detailed tables for the norm-scaling sweep on top of S. In this
experiment, the angular component is held fixed while the norm is multiplied by $\beta$. Thus,
changing $\beta$ does not change the target concept score; it only changes the radius of the steered
activation. This makes the sweep a direct test of whether the norm acts as an independent stability
lever.

\paragraph{PPL summary.}
Table~\ref{tab:app_t4_ppl} summarizes the mean PPL ratio for each $(\gamma,\beta)$ pair, together
with fold-level win counts. The monotone pattern at high $\gamma$ is the key result: once the
angular edit is large, increasing the norm reduces the PPL penalty.

\begin{table}[t]
\centering
\small
\caption{
Mean PPL ratio under the $\beta$ sweep. The best mean PPL for each $\gamma$ is highlighted in
bold. The lower block reports the number of folds in which each $\beta$ achieves the lowest PPL.
}
\label{tab:app_t4_ppl}
\begin{tabular}{lccccc}
\toprule
$\gamma$ & $\beta=0.9$ & $\beta=1.0$ (S) & $\beta=1.1$ & $\beta=1.2$ & Best \\
\midrule
0.1 & 1.10 & \textbf{1.10} & 1.11 & 1.12 & $\beta=1.0$ \\
0.3 & 1.76 & 1.71 & 1.69 & \textbf{1.69} & $\beta=1.2$ \\
0.5 & 9.82 & 7.98 & 6.88 & \textbf{6.21} & $\beta=1.2$ \\
0.7 & 262.5 & 151.8 & 107.2 & \textbf{83.5} & $\beta=1.2$ \\
\midrule
\multicolumn{6}{l}{\textit{Win counts: lowest PPL}} \\
0.1 & 26/70 & 25/70 & 9/70 & 10/70 & --- \\
0.3 & 5/70 & 10/70 & 15/70 & 40/70 & --- \\
0.5 & 2/70 & 0/70 & 6/70 & 62/70 & --- \\
0.7 & 0/70 & 0/70 & 0/70 & 70/70 & --- \\
\bottomrule
\end{tabular}
\end{table}

\paragraph{Task-metric summary.}
Table~\ref{tab:app_t4_task} reports the corresponding downstream task-metric changes. Compared
with PPL, task performance is much less sensitive to $\beta$: the spread across norm scales remains
small at each $\gamma$. This supports the interpretation that $\beta$ is primarily a stability knob,
not a semantic-control knob.

\begin{table}[t]
\centering
\small
\caption{
Downstream task-metric change under the $\beta$ sweep, in percentage points. ``Spread'' is the
maximum minus minimum over $\beta\in\{0.9,1.0,1.1,1.2\}$.
}
\label{tab:app_t4_task}
\begin{tabular}{lccccc}
\toprule
$\gamma$ & $\beta=0.9$ & $\beta=1.0$ (S) & $\beta=1.1$ & $\beta=1.2$ & Spread \\
\midrule
0.1 & +2.03 & +1.59 & +1.15 & +0.84 & 1.19pp \\
0.3 & +6.53 & +6.03 & +6.20 & +6.02 & 0.51pp \\
0.5 & +6.65 & +6.72 & +7.10 & +7.33 & 0.67pp \\
0.7 & +2.34 & +3.45 & +3.94 & +4.83 & 2.49pp \\
\bottomrule
\end{tabular}
\end{table}

\paragraph{Large-model sensitivity.}
Table~\ref{tab:app_t4_70b} isolates the 70B model. The larger model is more sensitive to strong
angular edits, producing larger PPL ratios at high $\gamma$, but the ordering over $\beta$ remains
the same. Larger norm scales still reduce PPL most strongly at high steering strengths.

\begin{table}[t]
\centering
\small
\caption{
70B-only PPL ratios under the $\beta$ sweep. The 70B model amplifies the PPL gap at high
$\gamma$, but the best $\beta$ ordering is unchanged.
}
\label{tab:app_t4_70b}
\begin{tabular}{lccccc}
\toprule
$\gamma$ & $\beta=0.9$ & $\beta=1.0$ (S) & $\beta=1.1$ & $\beta=1.2$ & Best \\
\midrule
0.1 & 1.09 & 1.08 & \textbf{1.08} & \textbf{1.08} & $\beta=1.1$--$1.2$ \\
0.3 & 2.13 & 1.96 & 1.85 & \textbf{1.78} & $\beta=1.2$ \\
0.5 & 27.6 & 19.2 & 14.2 & \textbf{11.2} & $\beta=1.2$ \\
0.7 & 585.8 & 453.1 & 354.6 & \textbf{281.6} & $\beta=1.2$ \\
\bottomrule
\end{tabular}
\end{table}

Overall, the $\beta$ sweep confirms that the radius is not merely a passive quantity. Once the
angular edit is fixed, changing the norm has little effect on the semantic task metric but a large
effect on generation stability. This strengthens the paper's two-parameter view of steering:
$\gamma$ controls the angular concept intervention, while $\beta$ controls the radial stability of
the resulting activation.

\section{Datasets and Data Sources}
\label{app:datasets}

This section summarizes the datasets used in our experiments. We use three groups of data:
concept datasets for direction construction and downstream steering evaluation, auxiliary
capability and language-modeling benchmarks, and unlabeled corpora for norm-variation
diagnostics.

\paragraph{Concept datasets.}
We evaluate steering on four concept datasets. TruthfulQA is used for truthfulness steering and
closed-form multiple-choice evaluation \citep{lin2022truthfulqa}. SST-2, derived from the Stanford
Sentiment Treebank, is used for sentiment steering \citep{socher2013recursive}. CivilComments is
used for toxicity and non-toxicity steering \citep{borkan2019nuanced}. IMDB is used as a second
sentiment dataset with longer movie-review inputs \citep{maas2011learning}. These datasets define
the contrastive concept directions and the task-specific downstream metrics reported in the main
experiments.

\paragraph{Auxiliary evaluation datasets.}
To measure whether steering degrades general model behavior, we evaluate perplexity on
WikiText-103 \citep{merity2016pointer}. We also evaluate general capability using MMLU, a
multi-task benchmark covering broad factual and reasoning domains \citep{hendrycks2021measuring}.
These auxiliary datasets are not used to construct steering directions; they are used only to measure
quality and capability retention under intervention.

\paragraph{Norm-variation corpora.}
For the norm-variation analysis, we use a heterogeneous set of corpora spanning web text,
instruction data, scientific writing, stories, question answering, toxicity comments, news,
biomedical text, and code. Specifically, we sample from OpenWebText \citep{gokaslan2019openwebtext},
Alpaca \citep{taori2023alpaca}, arXiv scientific papers \citep{cohan2018discourse}, WritingPrompts
\citep{fan2018hierarchical}, TruthfulQA \citep{lin2022truthfulqa}, Natural Questions
\citep{kwiatkowski2019natural}, CivilComments \citep{borkan2019nuanced}, CNN/DailyMail
\citep{hermann2015teaching,see2017get}, PubMedQA \citep{jin2019pubmedqa}, and CodeSearchNet
\citep{husain2019codesearchnet}. This mixture is intended to test whether the radial geometry of
hidden states is stable across content domains rather than being an artifact of a single dataset.

\paragraph{Dataset licenses.}
Table~\ref{tab:app_dataset_licenses} summarizes the licenses or usage terms associated with the
dataset distributions used in this work. Licenses vary across datasets and, in some cases, across
mirrors of the same dataset. We use all datasets only for research evaluation and do not redistribute
the datasets. For datasets whose original source does not specify a clear open-data license, we report
the relevant usage status conservatively and refer readers to the original source or distribution page.

\begin{table*}[t]
\centering
\small
\caption{
Dataset licenses or usage terms for the datasets used in our experiments. When a license depends
on the distribution mirror, we report the license of the distribution we rely on or note that the
license should be checked against the local source.
}
\label{tab:app_dataset_licenses}
\resizebox{\textwidth}{!}{
\begin{tabular}{llll}
\toprule
Dataset & Use in this work & License / usage terms & Source \\
\midrule
TruthfulQA
& Concept / evaluation
& Apache-2.0 for common HF distributions; check source distribution
& \citep{lin2022truthfulqa,truthfulqa_hf_license} \\

SST-2
& Concept / evaluation
& Distribution-dependent; common public mirrors list CC0 or CC-BY-4.0
& \citep{socher2013recursive,sst2_kaggle_license,sst2_zenodo_license} \\

CivilComments
& Concept / evaluation
& CC0-1.0 in the Hugging Face distribution
& \citep{borkan2019nuanced,civilcomments_hf_license} \\

IMDB
& Concept / evaluation
& Original Stanford release does not state a standard open-data license; used for research
& \citep{maas2011learning,imdb_stanford_license} \\

WikiText-103
& Perplexity evaluation
& CC BY-SA
& \citep{merity2016pointer,wikitext_hf_license} \\

MMLU
& Capability evaluation
& MIT in common public distributions
& \citep{hendrycks2021measuring,mmlu_hf_license} \\

OpenWebText
& Norm-variation corpus
& CC0 for the dataset packaging; underlying web text not owned by dataset authors
& \citep{gokaslan2019openwebtext,openwebtext_license} \\

Alpaca
& Norm-variation corpus
& CC BY-NC 4.0; research / non-commercial use
& \citep{taori2023alpaca,alpaca_license} \\

arXiv scientific papers
& Norm-variation corpus
& Distribution-dependent; license should be checked against the selected arXiv/paper source
& \citep{cohan2018discourse,scientific_papers_license} \\

WritingPrompts
& Norm-variation corpus
& MIT for the Hugging Face distribution used by common loaders
& \citep{fan2018hierarchical,writingprompts_hf_license} \\

Natural Questions
& Norm-variation corpus
& Creative Commons Share-Alike 3.0 on the Google download page
& \citep{kwiatkowski2019natural,nq_license} \\

CNN/DailyMail
& Norm-variation corpus
& Apache-2.0 in the Hugging Face distribution
& \citep{hermann2015teaching,see2017get,cnn_dailymail_hf_license} \\

PubMedQA
& Norm-variation corpus
& MIT
& \citep{jin2019pubmedqa,pubmedqa_license} \\

CodeSearchNet
& Norm-variation corpus
& Code/documentation MIT; source-code examples retain per-file upstream licenses
& \citep{husain2019codesearchnet,codesearchnet_license} \\
\bottomrule
\end{tabular}
}
\end{table*}

\section{Models and Licenses}
\label{app:models_licenses}

Table~\ref{tab:app_models_licenses} summarizes the model checkpoints used in our experiments,
together with their source families and licenses. We evaluate three model families: Llama
\citep{dubey2024llama3}, Qwen2.5 \citep{yang2024qwen25}, and Gemma 2
\citep{gemmateam2024gemma2}. All models are used only for research evaluation; we do not
redistribute model weights.

\begin{table*}[t]
\centering
\small
\caption{
Model checkpoints used in the experiments. Licenses are reported according to the corresponding
model cards or license pages.
}
\label{tab:app_models_licenses}
\resizebox{\textwidth}{!}{
\begin{tabular}{llll}
\toprule
Model checkpoint & Provider / family & Citation & License \\
\midrule
Llama-3.1-8B-Instruct
& Meta / Llama 3.1
& \citet{dubey2024llama3}
& Llama 3.1 Community License \citep{meta2024llama31license} \\

Llama-3.1-8B
& Meta / Llama 3.1
& \citet{dubey2024llama3}
& Llama 3.1 Community License \citep{meta2024llama31license} \\

Llama-3.1-70B-Instruct
& Meta / Llama 3.1
& \citet{dubey2024llama3}
& Llama 3.1 Community License \citep{meta2024llama31license} \\

Llama-3.2-1B-Instruct
& Meta / Llama 3.2
& \citet{dubey2024llama3}
& Llama 3.2 Community License \citep{meta2024llama32license} \\

Qwen2.5-7B-Instruct
& Alibaba / Qwen2.5
& \citet{yang2024qwen25}
& Apache-2.0 \citep{qwen2024qwen25license} \\

Qwen2.5-3B-Instruct
& Alibaba / Qwen2.5
& \citet{yang2024qwen25}
& Qwen Research License \citep{qwen2024qwen25researchlicense} \\

Gemma-2-9B-it
& Google / Gemma 2
& \citet{gemmateam2024gemma2}
& Gemma Terms of Use \citep{google2026gemmaterms} \\
\bottomrule
\end{tabular}
}
\end{table*}

The licenses differ in permissiveness. Qwen2.5-7B-Instruct is released under Apache-2.0, whereas
Qwen2.5-3B-Instruct is governed by the Qwen Research License. The Llama checkpoints are released
under Meta's Llama Community License, with separate license versions for Llama 3.1 and Llama
3.2. Gemma-2-9B-it is distributed under Google's Gemma Terms of Use. We report these licenses
for transparency and refer readers to the original model cards and license documents for the full
legal terms.

\end{document}